\documentclass[11pt,letterpaper]{article}
\usepackage{liveaigc_report}
\usepackage{mathtools,xspace,fix-cm}
\usepackage{tikz,pgfplots}
\pgfplotsset{compat=1.18}
\usetikzlibrary{arrows.meta,positioning,calc,fit,backgrounds,shapes.geometric,decorations.pathreplacing}
\usepackage[capitalize,nameinlink]{cleveref}
\definecolor{paperblue}{HTML}{235A83}
\definecolor{paperteal}{HTML}{237C70}
\definecolor{paperpurple}{HTML}{775A91}
\newcommand{\sg}{\operatorname{sg}}
\newcommand{\refset}{\mathcal R}
\newcommand{\hist}{\mathcal H}
\newcommand{\first}{\mathrm{first}}
\newcommand{\cont}{\mathrm{cont}}

\newcommand{\TaoMateResultLabel}{\makebox[0pt][l]{TaoMate-R1}\phantom{Temporal split (ours)}}

\newcommand{\MainResultsTable}{%
\begin{table*}[t]
\centering
\small
\setlength{\tabcolsep}{4pt}
\renewcommand{\arraystretch}{1.12}
\caption{\textbf{Multi-reference generation on 32 cases.} The upper block gives external systems; the lower block compares adaptation assignments on the same autoregressive backbone. TaoMate-R1 uses Temporal split. Values are case-wise means; bold indicates the best displayed mean within each block. Long-case external outputs concatenate instruction clips; autoregressive variants retain history.}
\label{tab:main_results}
\begin{tabular}{@{}l*{8}{c}@{}}
\toprule
\multirow{2}{*}{Method} & \multicolumn{2}{c}{Reference similarity} & Text & \multicolumn{2}{c}{Temporal similarity} & \multicolumn{3}{c}{VABench} \\
\cmidrule(lr){2-3}\cmidrule(lr){4-4}\cmidrule(lr){5-6}\cmidrule(l){7-9}
& CLIP $\uparrow$ & DINO $\uparrow$ & CLIP $\uparrow$ & CLIP $\uparrow$ & DINO $\uparrow$ & ViCLIP $\uparrow$ & Realism $\uparrow$ & QA $\uparrow$ \\
\midrule
MiniMax-H3 & $0.660$ & $0.355$ & $0.309$ & $0.934$ & $0.852$ & $\mathbf{0.218}$ & $4.50$ & $0.825$ \\
VACE & $\mathbf{0.680}$ & $\mathbf{0.390}$ & $\mathbf{0.323}$ & $\mathbf{0.946}$ & $\mathbf{0.893}$ & $0.217$ & $\mathbf{4.59}$ & $\mathbf{0.906}$ \\
SkyReels-V3 & $0.664$ & $0.372$ & $0.313$ & $0.919$ & $0.842$ & $0.208$ & $4.09$ & $0.575$ \\
Phantom & $0.664$ & $0.354$ & $0.321$ & $0.927$ & $0.866$ & $0.194$ & $4.19$ & $0.563$ \\
\midrule
Shared & $\mathbf{0.656}$ & $0.385$ & $\mathbf{0.306}$ & $0.951$ & $0.914$ & $\mathbf{0.215}$ & $4.63$ & $0.925$ \\
Operation split & $0.655$ & $0.388$ & $0.304$ & $0.953$ & $0.928$ & $0.213$ & $4.66$ & $0.938$ \\
\TaoMateResultLabel & $\mathbf{0.656}$ & $\mathbf{0.397}$ & $0.305$ & $\mathbf{0.957}$ & $\mathbf{0.934}$ & $0.206$ & $\mathbf{4.69}$ & $\mathbf{0.950}$ \\
\bottomrule
\end{tabular}
\end{table*}%
}

\newcommand{\BoundaryResultsTable}{%
\begin{table}[t]
\centering
\small
\setlength{\tabcolsep}{2.3pt}
\renewcommand{\arraystretch}{1.12}
\caption{\textbf{Instruction-boundary continuity.} Ten long sequences; each cell shows the mean above $\pm$ population standard deviation. TaoMate-R1 uses Temporal split. CLIP/DINO compare adjacent boundary frames; Luma is the absolute change in mean luminance. Bold indicates the best mean.}
\label{tab:boundary_results}
\begin{tabular}{@{}lccc@{}}
\toprule
Generator & CLIP $\uparrow$ & DINO $\uparrow$ & Luma $\downarrow$ \\
\midrule
Shared & \shortstack{$0.979$\\$\pm\,0.00737$} & \shortstack{$0.978$\\$\pm\,0.00585$} & \shortstack{$1.86$\\$\pm\,0.626$} \\[3pt]
Operation split & \shortstack{$0.979$\\$\pm\,0.00751$} & \shortstack{$0.973$\\$\pm\,0.0160$} & \shortstack{$1.96$\\$\pm\,0.456$} \\[3pt]
\TaoMateResultLabel & \shortstack{$\mathbf{0.991}$\\$\pm\,0.00268$} & \shortstack{$\mathbf{0.990}$\\$\pm\,0.00412$} & \shortstack{$\mathbf{0.613}$\\$\pm\,0.103$} \\
\bottomrule
\end{tabular}
\end{table}%
}

\newcommand{\SubgroupResultsTable}{%
\begin{table}[t]
\centering
\small
\setlength{\tabcolsep}{3pt}
\renewcommand{\arraystretch}{1.12}
\caption{\textbf{Specialization across interaction settings.} Short single-instruction clips, long changing-instruction streams, and public reference compositions. TaoMate-R1 uses Temporal split. Scores are case-wise means; higher is better. Bold indicates the best displayed mean within each group.}
\label{tab:subgroup_results}
\begin{tabular}{@{}lcccc@{}}
\toprule
Generator & DINO$_{\rm ref}$ & CLIP$_{\rm temp}$ & DINO$_{\rm temp}$ & QA \\
\midrule
\multicolumn{5}{@{}l}{\textit{Short (10 cases)}} \\
Shared & $0.525$ & $0.967$ & $0.962$ & $0.820$ \\
Operation split & $0.531$ & $0.969$ & $0.972$ & $0.900$ \\
\TaoMateResultLabel & $\mathbf{0.545}$ & $\mathbf{0.970}$ & $\mathbf{0.975}$ & $0.880$ \\
\midrule
\multicolumn{5}{@{}l}{\textit{Long (10 cases)}} \\
Shared & $0.469$ & $0.943$ & $0.891$ & $\mathbf{1.00}$ \\
Operation split & $\mathbf{0.486}$ & $0.948$ & $0.914$ & $\mathbf{1.00}$ \\
\TaoMateResultLabel & $0.478$ & $\mathbf{0.956}$ & $\mathbf{0.921}$ & $0.980$ \\
\midrule
\multicolumn{5}{@{}l}{\textit{Public (12 cases)}} \\
Shared & $0.200$ & $0.944$ & $0.892$ & $0.950$ \\
Operation split & $0.187$ & $0.944$ & $0.903$ & $0.917$ \\
\TaoMateResultLabel & $\mathbf{0.207}$ & $\mathbf{0.948}$ & $\mathbf{0.911}$ & $\mathbf{0.983}$ \\
\bottomrule
\end{tabular}
\end{table}%
}

\title{TaoMate-R1: Temporal Role Specialization for Streaming Multi-Reference Video Generation}
\reportauthors{Qijun Gan\textsuperscript{*}, Chenwei Zhang\textsuperscript{*}, Xinyang Li,
Meiguang Jin\textsuperscript{\textdagger}, Junfeng Ma}
\authornote{\textsuperscript{*}Equal contribution.\quad \textsuperscript{\textdagger}Corresponding author.}
\affiliation{Alibaba Group\\[0.3em]
\texttt{\{ganqijun.gqj, meiguang.jmg\}@taobao.com}}
\teamname{}
\shorttitle{TaoMate-R1: Streaming Multi-Reference Video Generation}
\reportlabel{Technical Report}
\reportdate{November 2026}
\leftlogo[height=0.9cm]{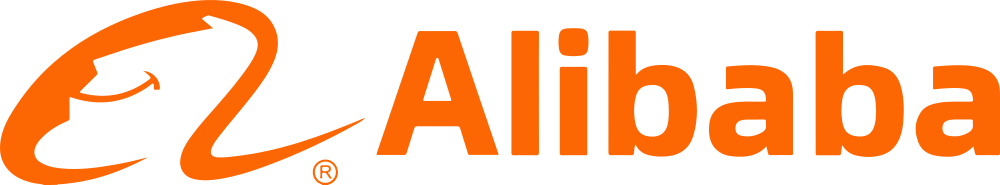}
\rightlogo[height=0.9cm,trim=436bp 638bp 306bp 480bp,clip]{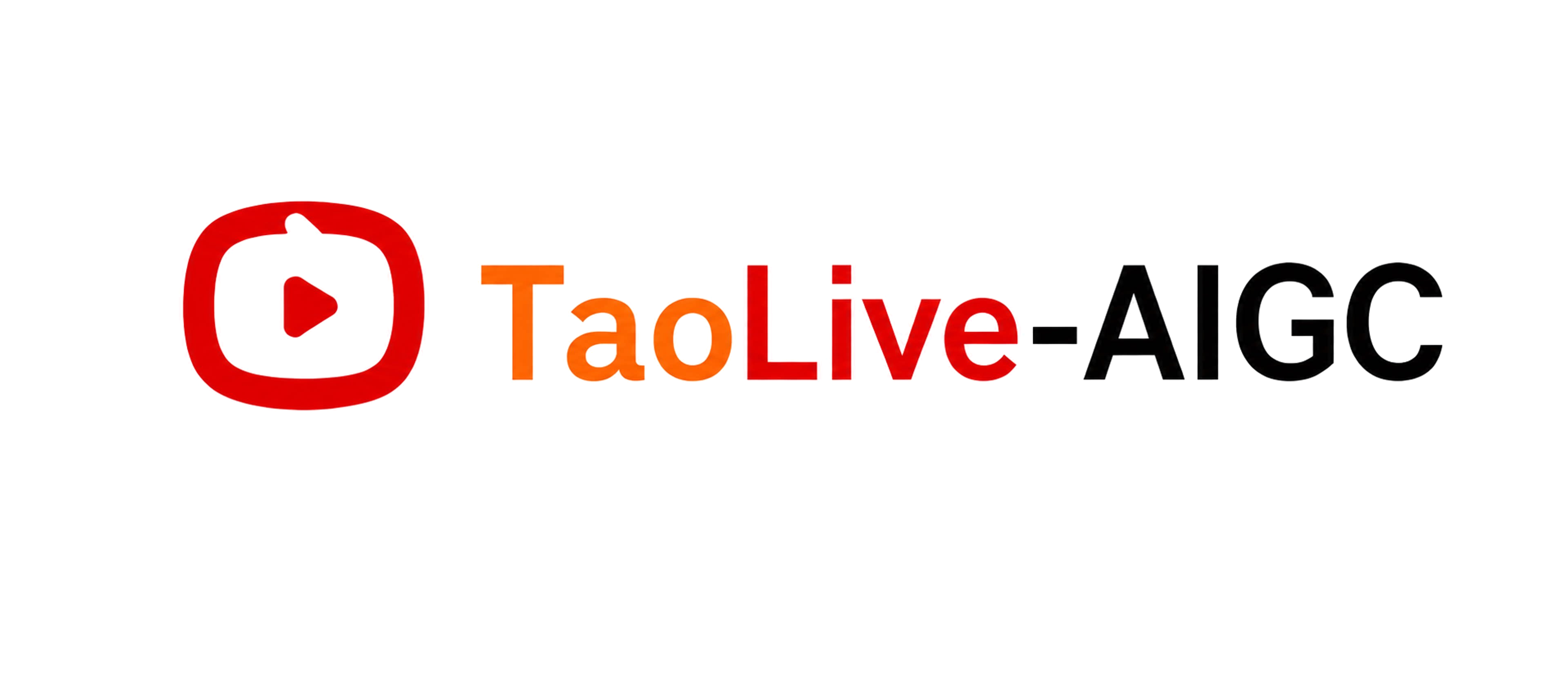}
\hypersetup{
  pdftitle={TaoMate-R1: Temporal Role Specialization for Streaming Multi-Reference Video Generation},
  pdfauthor={Qijun Gan, Chenwei Zhang, Xinyang Li, Meiguang Jin, Junfeng Ma},
  pdfsubject={TaoMate-R1 for streaming multi-reference video generation through temporal role specialization},
  pdfkeywords={TaoMate-R1, streaming video generation, multi-reference generation, temporal role specialization}
}

\begin{document}
\maketitle
\begin{abstract}
Streaming multi-reference video generation establishes a scene from specified people and objects, then continues it under changing instructions.
We introduce TaoMate-R1, a streaming multi-reference generator that organizes adaptation parameters through temporal role specialization.
It assigns separate low-rank adapters to scene establishment and continuation on a frozen backbone.
Each adapter performs both denoising and context writing for its role; Self Gradient Forcing jointly trains the roles through the history they construct.
This allocation contrasts with operation-based specialization, which separates denoising from context writing across all chunks.
A fixed-replay gradient analysis characterizes the update directions admitted by the two sharing schemes.
On 32 multi-reference cases, we compare the schemes at the same two-adapter budget.
Across ten approximately two-minute streams with changing instructions and shared teacher-audio conditioning, TaoMate-R1 reduces the mean boundary luminance jump from 1.96 to 0.613, with higher boundary feature similarity.
It also achieves higher mean CLIP and DINO temporal similarity across short interactions, long streams, and public reference compositions.
These results support grouping adaptation parameters by scene establishment and continuation for streaming multi-reference generation. Our code and model are released at \url{https://github.com/TaoLiveAIGC/TaoMate-H3}. 
\end{abstract}

\begin{figure*}[t]
\centering
\resizebox{\textwidth}{!}{\input{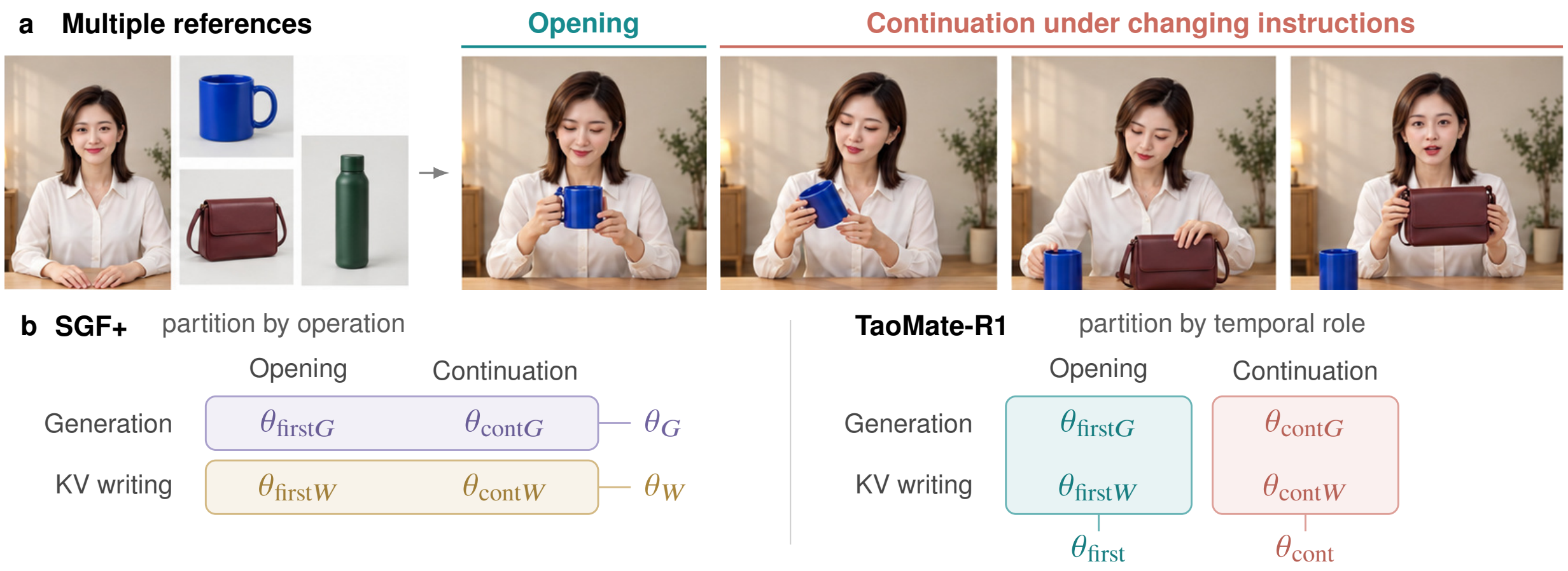}}
\par\vspace{2pt}
\resizebox{\textwidth}{!}{
\begin{tikzpicture}[x=1cm,y=1cm,
  font=\sffamily\fontsize{8.5}{10}\selectfont,
  coordinate label/.style={font=\fontsize{10}{11}\selectfont,inner sep=0},
  adapter label/.style={font=\fontsize{10}{11}\selectfont,inner sep=0},
  partition/.style={rounded corners=3pt,line width=.55pt},
  guide/.style={line width=.35pt,draw=black!20}]
\definecolor{specgen}{HTML}{7C6EB0}
\definecolor{specwrite}{HTML}{BC9442}
\definecolor{specfirst}{HTML}{188B8E}
\definecolor{speccont}{HTML}{CB6C60}
\path[use as bounding box] (0,0) rectangle (16.5,2.66);

\node[anchor=west,font=\sffamily\bfseries\fontsize{9}{11}\selectfont]
  at (.04,2.45) {b};
\node[anchor=west,font=\sffamily\bfseries\fontsize{9}{11}\selectfont]
  at (.40,2.45) {SGF+};
\node[anchor=west,text=black!65,font=\sffamily\fontsize{8.5}{10}\selectfont]
  at (1.52,2.45) {partition by operation};
\node[anchor=west,font=\sffamily\bfseries\fontsize{9}{11}\selectfont]
  at (8.76,2.45) {TaoMate-R1};
\node[anchor=west,text=black!65,font=\sffamily\fontsize{8.5}{10}\selectfont]
  at (11.10,2.45) {partition by temporal role};
\draw[guide] (8.22,.16) -- (8.22,2.51);

\foreach \origin in {0,8.36}{
  \begin{scope}[shift={(\origin,0)}]
    \node[text=black!72] at (3.08,1.98) {Opening};
    \node[text=black!72] at (5.24,1.98) {Continuation};
    \node[anchor=east,text=black!72] at (1.90,1.43) {Generation};
    \node[anchor=east,text=black!72] at (1.90,.76) {KV writing};
  \end{scope}
}

\draw[partition,draw=specgen!65,fill=specgen!10]
  (2.11,1.15) rectangle (6.21,1.71);
\draw[partition,draw=specwrite!65,fill=specwrite!11]
  (2.11,.48) rectangle (6.21,1.04);
\draw[specgen!65,line width=.55pt] (6.21,1.43) -- (6.55,1.43);
\draw[specwrite!65,line width=.55pt] (6.21,.76) -- (6.55,.76);
\node[adapter label,text=specgen!90!black,anchor=west] at (6.68,1.43) {$\theta_G$};
\node[adapter label,text=specwrite!90!black,anchor=west] at (6.68,.76) {$\theta_W$};
\node[coordinate label,text=specgen!85!black] at (3.08,1.43) {$\theta_{\first G}$};
\node[coordinate label,text=specgen!85!black] at (5.24,1.43) {$\theta_{\cont G}$};
\node[coordinate label,text=specwrite!85!black] at (3.08,.76) {$\theta_{\first W}$};
\node[coordinate label,text=specwrite!85!black] at (5.24,.76) {$\theta_{\cont W}$};

\begin{scope}[shift={(8.36,0)}]
  \draw[partition,draw=specfirst!65,fill=specfirst!10]
    (2.11,.48) rectangle (4.05,1.71);
  \draw[partition,draw=speccont!65,fill=speccont!10]
    (4.27,.48) rectangle (6.21,1.71);
  \node[coordinate label,text=specfirst!90!black] at (3.08,1.43) {$\theta_{\first G}$};
  \node[coordinate label,text=specfirst!90!black] at (3.08,.76) {$\theta_{\first W}$};
  \node[coordinate label,text=speccont!90!black] at (5.24,1.43) {$\theta_{\cont G}$};
  \node[coordinate label,text=speccont!90!black] at (5.24,.76) {$\theta_{\cont W}$};
  \draw[specfirst!65,line width=.55pt] (3.08,.48) -- (3.08,.30);
  \draw[speccont!65,line width=.55pt] (5.24,.48) -- (5.24,.30);
  \node[adapter label,text=specfirst!90!black] at (3.08,.12) {$\theta_\first$};
  \node[adapter label,text=speccont!90!black] at (5.24,.12) {$\theta_\cont$};
\end{scope}
\end{tikzpicture}}
\caption{\textbf{Streaming multi-reference video generation.}
(a) Multiple references guide scene establishment and continuation under changing instructions (conceptual illustration).
(b) Each colored group shares one adapter: SGF+ groups computations by operation; TaoMate-R1 groups them by temporal role, sharing generation and context-writing parameters within each role.}
\label{fig:overview}
\end{figure*}

\section{Introduction}
\label{sec:intro}

Multi-reference video generation specifies the people and objects in a scene through images.
Models such as Phantom, VACE, and MiniMax-H3 compose these visual conditions into video clips~\citep{phantom,vace,minimaxh3}.
Streaming extends the task to an evolving scene: a presenter introduces a cup, puts it away, and picks up a bag as new instructions arrive.
The desired behavior combines change with persistence.
The current instruction determines the interaction, while the generated past determines its starting state.

The opening and subsequent chunks therefore solve different conditional tasks.
At the opening, the generator chooses how reference appearances are realized through pose, placement, and initial motion.
During continuation, those realized choices become part of the conditioning: a hand holding a cup constrains where the next action begins.
This asymmetry motivates studying whether these tasks should receive independent adaptation.

Autoregressive diffusion and student-rollout training make generated history available during learning~\citep{causvid,selfforcing}.
Self Gradient Forcing (SGF) also lets later predictions supervise the computations that write this history~\citep{sgf}.
SGF+ separates denoising and context-writing parameters to accommodate differences between their gradient contributions~\citep{sgfplus}.
This is an operation-based partition: each operation shares parameters across the entire stream.
We investigate a temporal-role partition at the same two-adapter budget, grouping generation and writing by whether they establish or continue a scene.
The distinction concerns which computations learn together (\cref{fig:overview}).

We introduce \emph{TaoMate-R1}, which implements \emph{temporal role specialization} with an opening adapter at absolute chunk zero and a continuation adapter for every later chunk.
Within each role, denoising and context writing share the same low-rank parameters on a frozen backbone.
Instruction changes update the continuation conditions while preserving its adapter and generated history.
We train both roles with SGF: historical-context gradients return to the role that wrote the corresponding representation.
The opening role thus learns both the first chunk and its persistent video anchor; the continuation role learns subsequent predictions and recent audiovisual context.

We characterize this design through a local analysis of parameter sharing.
At a common initialization and fixed replay, the two partitions preserve the same common gradient component but admit different specialization directions.
The analysis also isolates continuation-only windows, where the comparison depends on alignment between continuation's generation and writing gradients.
These relations describe the optimization consequences of tying parameters and provide a precise basis for comparing the two assignments.

Our experiments compare TaoMate-R1's Temporal split with Shared and Operation split on the same H3 backbone.
Across 32 cases, TaoMate-R1 achieves higher mean CLIP and DINO temporal similarity than Operation split on short interactions, long instruction-driven streams, and public reference compositions.
The clearest difference occurs at instruction boundaries in ten approximately two-minute streams: TaoMate-R1 has a mean luminance jump of 0.613, compared with 1.96 for Operation split, and higher adjacent-frame feature similarity.
We examine these differences with case-paired statistics and background-appearance visualizations, then place the generator's results alongside external reference-conditioned systems.
The contributions are:
\begin{itemize}
    \item TaoMate-R1, a streaming multi-reference generator that assigns independent adapters to scene establishment and continuation, shares denoising and context writing within each role, and trains them jointly through SGF.
    \item A local gradient characterization of temporal and operation-based sharing, including the dependence on generation--writing alignment in continuation-only training windows.
    \item A comparison across 32 multi-reference cases showing higher mean temporal similarity across three interaction settings and reduced instruction-boundary discontinuities on long streams.
\end{itemize}

\FloatBarrier
\section{Related Work}
\label{sec:related}

\paragraph{Multi-Reference Generation and Control.}
Reference images specify visual attributes and associate them with entities.
ConsisID separates facial structure from identity detail~\citep{consisid}; Video Alchemist binds reference features to entity words~\citep{videoalchemist}; and Movie Weaver combines prompt anchors with reference-order embeddings~\citep{movieweaver}.
Phantom aligns visual and textual reference representations~\citep{phantom}, VACE integrates image, video, and mask conditions~\citep{vace}, and MiniMax-H3 supports multi-reference audiovisual generation~\citep{minimaxh3}.
Spatiotemporal control specifies where and when these entities appear: DreamVideo-2 uses bounding-box trajectories~\citep{dreamvideo2}, AlcheMinT uses interval-dependent rotary embeddings~\citep{alchemint}, and MultiShotMaster combines shot-aware encoding with reference grounding~\citep{multishotmaster}.
For interactive streams, FlowAct-R2 aligns evolving conditions with planned behavior~\citep{flowactr2}, while ReferStream introduces arriving subjects through contextualized reference latents and cache injection~\citep{referstream}.
These methods organize visual conditions; our study concerns the adaptation parameters used to realize them across an ongoing scene.

\paragraph{Causal Generation and Generated-History Training.}
Diffusion Forcing uses token-wise noise levels for causal generation~\citep{diffusionforcing}; CausVid distills a bidirectional teacher into a few-step causal model~\citep{causvid}; and SkyReels-V2 extends videos with nondecreasing frame-wise noise schedules~\citep{skyreelsv2}.
History retention supplies persistent appearance and recent motion: StreamingT2V combines initial-chunk cues with preceding-chunk features~\citep{streamingt2v}, FramePack compresses context according to historical importance~\citep{framepack}, and LongLive combines a frame sink with local attention~\citep{longlive}.
LongLive-RAG retrieves earlier latents~\citep{longliverag}, while FilmWeaver separates cross-shot keyframes from recent within-shot memory~\citep{filmweaver}.
Training with generated predecessors addresses the context distribution encountered at inference, from Scheduled Sampling~\citep{scheduledsampling} to Self Forcing's student rollouts~\citep{selfforcing} and Rolling Forcing's jointly denoised windows~\citep{rollingforcing}.
SGF additionally propagates supervision through historical context writers~\citep{sgf}.
We adopt this training framework and study parameter sharing under a fixed history-retention policy.

\paragraph{Parameter Specialization.}
Specialization allocates independent parameters to computations with different learning demands.
DreamVideo composes identity and motion adapters~\citep{dreamvideo}, while eDiff-I and Wan2.2 specialize denoisers by diffusion-noise interval~\citep{ediffi,wan22}.
SGF+ separates denoising and context-writing parameters to address gradient differences between these operations~\citep{sgfplus}.
TaoMate-R1 groups the same operations by temporal role, using independent low-rank adapters~\citep{lora} for scene establishment and continuation.
Our comparison with operation-based sharing holds the two-adapter budget fixed and examines the resulting generation and continuity measurements.

\FloatBarrier
\section{TaoMate-R1}
\label{sec:method}
\begin{figure*}[t]
\centering
\resizebox{\textwidth}{!}{\input{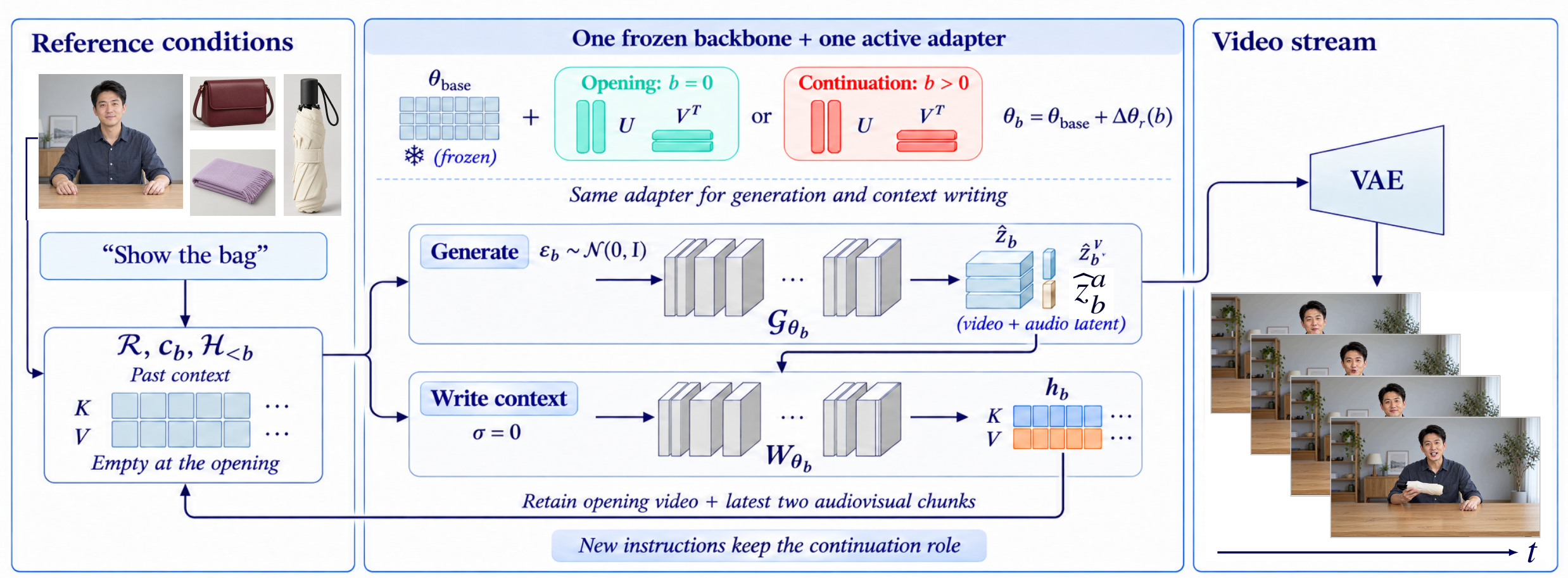}}
\caption{\textbf{TaoMate-R1: generation and context writing through temporal role specialization.}
The absolute chunk index selects an adapter for both denoising and clean-context writing.
The resulting audiovisual keys and values condition subsequent chunks; video latents are decoded for display.
The reference images and four stacked output frames come from an evaluated product-interaction sequence.}
\label{fig:architecture}
\end{figure*}

TaoMate-R1 is a streaming multi-reference generator built on temporal role specialization: it groups generation and context writing by their role in the stream (\cref{fig:architecture}).

\subsection{Preliminaries}
\label{sec:preliminaries}
\paragraph{Generation and Context Writing.}
A streaming diffusion model alternates between generating a chunk and encoding it as history.
For external conditions $q_b$ and retained keys and values $\hist_{<b}$, a denoising trajectory $G_\theta$ maps noise $\epsilon_b$ to a clean latent; a clean-time execution $W_\theta$ writes context:
\begin{equation}
\begin{aligned}
\widehat z_b &=G_\theta(\epsilon_b,q_b,\hist_{<b}),\\
h_b &=W_\theta(\widehat z_b,q_b,\hist_{<b};0).
\end{aligned}
\label{eq:prelim_stream}
\end{equation}
The retained part of $h_b$ conditions subsequent predictions.

\paragraph{Generated-History Supervision.}
SGF trains historical representations on student-generated rollouts~\citep{sgf}.
Recorded historical latent inputs remain fixed during differentiation, while later losses propagate through their KV-writing computations.
Generation and writing therefore contribute separate paths to the same objective.
Sharing their parameters sums these contributions into one update; independent parameters allocate them to separate updates.
Our method specifies this sharing relation for the opening and continuation of a stream.

\subsection{Problem Formulation}
\label{sec:stream}
Let $\refset_b$ denote the reference images active at chunk $b$, $c_b$ its instruction, and $z_b=(z_b^v,z_b^a)$ its audiovisual latents.
These specify $q_b=(\refset_b,c_b)$ in \cref{eq:prelim_stream}.
The retained history $\hist_{<b}$ consists of layerwise audiovisual KVs from preceding chunks, with $\hist_{<0}=\varnothing$ at the opening.
We parameterize a stream of $B$ chunks as
\begin{equation}
\begin{aligned}
&p(z_{0:B-1}\mid\refset_{0:B-1},c_{0:B-1})\\
&\quad=p_{\theta_\first}(z_0\mid\refset_0,c_0)
\prod_{b=1}^{B-1}
p_{\theta_\cont}(z_b\mid\refset_b,c_b,\hist_{<b}).
\end{aligned}
\label{eq:factorization}
\end{equation}
Establishment occurs once, at absolute chunk $b=0$.
Changes in instructions or active references become conditions for continuation within the existing scene.

The references and generated history provide different kinds of information.
Reference images identify entities and their appearances; history records how those entities have been realized together in the current scene.
At the opening, the model establishes their spatial arrangement, poses, and interaction state from the supplied conditions.
Continuation inherits this realization and predicts its evolution under the current instruction.
For example, presenting a new object changes the active references while the presenter and room remain part of the ongoing scene.
The temporal role thus follows the stream's established state: changing an instruction updates the interaction goal while retaining the history from which the next action begins.

The long-stream evaluation additionally supplies cached audio states and removal-segment end-frame conditions.

\subsection{Temporal Role Specialization}
\label{sec:specialization}
The opening and continuation adapters use independent LoRA updates~\citep{lora} on a shared frozen backbone:
\begin{equation}
r(b)=
\begin{cases}
\first,& b=0,\\
\cont,& b>0,
\end{cases}
\qquad
\theta_b=\theta_{\mathrm{base}}+\Delta\theta_{r(b)}.
\label{eq:routing}
\end{equation}
Both inherit the LoRA weights of the same trained autoregressive generator and are optimized jointly.
Their weights, AdamW moments, and exponential moving averages are copied into independent states.
For each adapted layer, $\Theta_r=\Theta_0+B_rA_r$, with fixed LoRA scaling absorbed into the factors.
Both adapters use rank and scaling 128 on the attention QKV/output and feed-forward projections, covering 208 linear layers.

The selected $\theta_b$ governs both $G$ and $W$ in \cref{eq:prelim_stream}.
Thus, a temporal role adapts its synthesis mapping together with the context representation it leaves for future predictions.

This grouping specifies which learning signals shape the same adaptation.
For establishment, generation realizes the references in an initial scene and writing represents that scene for subsequent predictions.
For continuation, generation evolves the interaction and writing updates the context available to the next chunk.
Within-role sharing combines these contributions according to their conditional task.
Operation-based specialization instead combines opening and continuation signals within each operation.
The same generation and writing calls are therefore grouped along different axes: temporal role in TaoMate-R1 and computational operation in SGF+.

At inference, the history retains the opening video KVs as a persistent appearance anchor and the latest two audiovisual chunks' KVs as recent interaction context.
For $b>0$,
\begin{equation}
\hist_{<b}=\{h_0^v\}\ \cup\
\bigcup_{j=\max(0,b-2)}^{b-1}\{h_j^v,h_j^a\}.
\label{eq:retention}
\end{equation}
Overlapping entries are counted once; opening audio remains only in the recent window.
Reference conditions remain available independently of this history, and temporal positions remain absolute.
Thus, the opening role writes the persistent anchor, while continuation supplies the rolling interaction context.
All three parameter-sharing configurations use this retention policy.

\subsection{Joint Training with Context Gradients}
\label{sec:training}
\paragraph{Student Rollouts and Distribution Matching.}
We train TaoMate-R1's two roles using student rollouts~\citep{selfforcing} and SGF's context gradients~\citep{sgf}.
Let $\mathcal I=\{s,\ldots,e\}$ denote a sampled training window, with its preceding prefix held fixed as conditioning.
We request five or ten chunks with probabilities 0.8 and 0.2, truncated by source length.
The prefix is encoded from cached clean teacher latents; references and text remain fixed within an example.
Nominal opening and continuation chunks contain 39 and 34 frames at 24\,fps.
A gradient-free rollout uses one initial-to-clean evaluation per chunk and records the prediction noise, generated latents, context, and actual writer input $\widetilde z_b$, including the training context perturbation.
Both computations use the adapter selected by the absolute chunk index.

Distribution-matching distillation~\citep{dmd2} supervises the generated audiovisual content.
A frozen real-score model and a learned fake-score model evaluate the same corrupted student sequence, producing a detached direction $d_b$.
Let $\widehat z_{b,\mathrm{real}}^m$ and $\widehat z_{b,\mathrm{fake}}^m$ be their clean estimates for modality $m$.
We use the normalized direction
\begin{equation}
d_b^m=\sg\!\left[
\frac{\widehat z_{b,\mathrm{fake}}^m-\widehat z_{b,\mathrm{real}}^m}
{\max\!\left(\operatorname{mean}|\widehat z_b^m-\widehat z_{b,\mathrm{real}}^m|,\varepsilon\right)}
\right].
\label{eq:score_direction}
\end{equation}
We set $\varepsilon=10^{-8}$.
For a prediction $\widehat z_b$ replayed with its recorded inputs, we optimize
\begin{equation}
\mathcal L_b=\frac12\sum_{m\in\{v,a\}}\lambda_m
\left\|\widehat z_b^m-\sg(\widehat z_b^m-d_b^m)\right\|_{\mathrm{mean}}^2.
\label{eq:main_surrogate}
\end{equation}
Here $\sg$ stops gradients, $\lambda_v=1$, $\lambda_a=0.1$, and the squared norm averages over that modality's latent elements.
The window objective is $\mathcal L=|\mathcal I|^{-1}\sum_{b\in\mathcal I}\mathcal L_b$.
Both roles share a single fake-score model, trained on detached student outputs.
The fake score receives velocity-regression updates every iteration; generator updates occur every fifth iteration.

\paragraph{Historical Credit Assignment.}
\label{sec:history_training}
Replay losses accumulate context gradients on shared historical KV variables.
We realize SGF's context-gradient reconstruction by differentiating each historical writer once in reverse temporal order, with its recorded latent input held fixed.
The accumulated gradient returns to the adapter selected by that writer's absolute chunk index and propagates through its preceding KVs.
This assigns both synthesis and historical credit to the appropriate temporal role:
\begin{equation}
\nabla_{\Delta\theta_r}\mathcal L
=
\sum_{b\in\mathcal I_r}g_b^{\mathrm{gen}}
+
\sum_{j\in\mathcal J_r}
\left(\frac{\partial h_j}{\partial\Delta\theta_r}\right)^{\!\top}
\bar h_j.
\label{eq:role_credit}
\end{equation}
Here $\mathcal I_r$ and $\mathcal J_r$ index role $r$'s supervised predictions and differentiable historical writers within the effective window.
The direct gradient $g_b^{\mathrm{gen}}$ is taken through the current denoiser from $\mathcal L_b/|\mathcal I|$.
The writer Jacobian holds its recorded latent input and incoming KVs fixed; $\bar h_j$ includes all downstream prediction and writer paths under the window-averaged objective.

A historical representation influences the objective through direct reads by later denoisers and through context written by intervening chunks.
The reverse pass accumulates these routes before differentiating the corresponding writer.
For a recorded writer input, the context computation and its contribution to an earlier representation are
\begin{align}
h_j&=W_{\theta_j}\bigl(\sg(\widetilde z_j),q_j,\hist_{<j};0\bigr),
\label{eq:writer_graph}\\
\bar h_i&\mathrel{+}=
\left(\frac{\partial h_j}{\partial h_i}\right)^{\!\top}\bar h_j,
\qquad s\le i<j.
\label{eq:history_grad}
\end{align}
The same accumulated signal supplies the writer term in \cref{eq:role_credit}.
Its parameter update belongs to the role that produced the representation, connecting independently adapted opening and continuation computations through their shared generated history.

When the effective window includes the opening, continuation losses supervise its context representation through $\Delta\theta_\first$.
For $s>0$, the opening is fixed conditioning and receives no update.
Algorithm~\ref{alg:training} summarizes the rollout and reverse context-gradient pass.
\begin{algorithm}[t]
\caption{TaoMate-R1: temporal role specialization with shared-history supervision}
\label{alg:training}
\small
\begin{algorithmic}[1]
\Require Window $\mathcal I=\{s,\ldots,e\}$, fixed prefix, references $\refset_b$, instructions $c_b$, adapters $\theta_\first,\theta_\cont$
\State $r(b)\gets\first$ if $b=0$, otherwise $\cont$
\Statex \textbf{Pass 1: self-generated rollout, without gradients}
\For{$b=s,\ldots,e$}
  \State $\widehat z_b\gets G_{\theta_{r(b)}}(\epsilon_b,\refset_b,c_b,\hist_{<b})$
  \State $h_b\gets W_{\theta_{r(b)}}(\widetilde z_b,\refset_b,c_b,\hist_{<b};0)$
  \State Record inputs and context; advance history with $h_b$
\EndFor
\Statex The shared fake score learns from detached samples each iteration.
\Statex \textbf{Pass 2: scheduled generator update}
\State Expose window KVs as shared gradient leaves; keep the prefix fixed
\State Initialize role gradients $g_r\gets0$ and KV cotangents $\bar h_j\gets0$
\For{$b=s,\ldots,e$}
  \State Replay current $G_{\theta_{r(b)}}$ with recorded inputs
  \State Form $\mathcal L_b$ using detached score direction (\ref{eq:main_surrogate})
  \State Accumulate $\nabla(\mathcal L_b/|\mathcal I|)$ into $g_{r(b)}$ and shared $\bar h_j$
\EndFor
\For{$j=e-1,\ldots,s$}
  \State Reconstruct $W_{\theta_{r(j)}}$ from $\sg(\widetilde z_j)$ and the corresponding shared KV leaves
  \State Apply VJP with $\bar h_j$; add to $g_{r(j)}$ and earlier KV cotangents
\EndFor
\State Update each supervised adapter with its accumulated $g_r$
\end{algorithmic}
\end{algorithm}

\paragraph{Optimization and Inference.}
We use AdamW with generator/fake-score learning rates $5\times10^{-5}/10^{-5}$, $\beta=(0,0.999)$, weight decay 0.01, gradient clipping at 1, global batch 16, and seed 42.
Training perturbs noninitial writer inputs up to noise level 0.15; writing still uses clean time, and replay retains the sampled perturbation.
Inference uses unperturbed context and three Euler evaluations per chunk.
With $u_{b,0}=\epsilon_b$ and velocity prediction $v_{\theta_b}^m$, each evaluation updates
\begin{equation}
\begin{aligned}
u_{b,k+1}^m={}&u_{b,k}^m+(\sigma_k^m-\sigma_{k+1}^m)\\
&\qquad\cdot v_{\theta_b}^m(u_{b,k},q_b,\hist_{<b}).
\end{aligned}
\label{eq:euler}
\end{equation}
For $u_i=1-i/49$, the noise schedule is $\sigma_i^m=\gamma_m u_i/[1+(\gamma_m-1)u_i]$, with $\gamma_v=12$, $\gamma_a=3$ and retained indices $(0,16,33,49)$.
The selected role remains active throughout denoising and clean-context writing.

\subsection{Local Gradient Analysis of Parameter Sharing}
\label{sec:axis}
To compare the partitions in Figure~\ref{fig:overview}(b), introduce four virtual adapter copies, one per temporal-role--operation pair:
$x=(\theta_{\first G},\theta_{\first W},\theta_{\cont G},\theta_{\cont W})$.
Temporal role specialization ties generation and writing within each temporal role; SGF+'s operation split ties opening and continuation within each operation~\citep{sgfplus}.

Consider a fixed replay at the common adapter initialization, holding the sampled rollout, historical latent inputs, prediction noise, and distribution-matching targets fixed.
Let $g_{rG}$ and $g_{rW}$ denote the gradient contributions through generation and writing for temporal role $r$.
The contrasts retained by the two assignments are
\begin{equation}
\begin{aligned}
v_T&=(g_{\first G}+g_{\first W})-(g_{\cont G}+g_{\cont W}),\\
v_O&=(g_{\first G}+g_{\cont G})-(g_{\first W}+g_{\cont W}).
\end{aligned}
\label{eq:contrasts}
\end{equation}
For a fixed norm $\rho$ of the four-block parameter update, let $D_T$ and $D_O$ be the largest first-order loss decreases admitted by temporal and operation sharing.
Orthogonal projection gives
\begin{equation}
D_T^2-D_O^2=\frac{\rho^2}{4}
\bigl(\|v_T\|^2-\|v_O\|^2\bigr).
\label{eq:axis_advantage}
\end{equation}
Both assignments retain the common gradient while preserving a different specialization direction.
Temporal role specialization admits the larger local decrease when the establishment--continuation contrast exceeds the denoising--writing contrast in norm.
For a window starting after chunk zero, the opening belongs to the fixed prefix, so $g_{\first G}=g_{\first W}=0$ and
\begin{equation}
D_T^2-D_O^2=\rho^2\langle g_{\cont G},g_{\cont W}\rangle.
\label{eq:continuation_alignment}
\end{equation}
In this case, aligned generation and writing gradients favor sharing within continuation; opposing gradients favor operation separation.
The relations characterize a fixed replay's local update geometry; our experiments compare the resulting trained generators.

\FloatBarrier
\section{Experiments}
\label{sec:experiments}
We evaluate TaoMate-R1 through parameter-sharing comparisons on a common streaming backbone, followed by system-level generation results.

\subsection{Experimental Setup}
\paragraph{Cases and Parameter-Sharing Configurations.}
We evaluate 32 cases: ten short person--object interactions, ten approximately two-minute streams with twenty successive instructions each, and twelve public reference compositions from MultiRef-Compass~\citep{multirefcompass}.
Short and public clips are approximately ten seconds long.
On the H3 backbone, \emph{Shared} uses one rank-128 adapter for generation and writing across all chunks.
\emph{Operation split} uses two rank-128 adapters, separating generation from writing following SGF+.
TaoMate-R1 implements \emph{Temporal split} with two rank-128 adapters, separating opening from continuation.
The two splits have the same adapter budget; their grouping of computations is the variable of interest.
All three configurations use the same training cohort, reference inputs, instructions, history-retention policy, and three-evaluation Euler solver.
Short and public cases generate native audio; long cases follow cached teacher-audio trajectories and use end-frame conditions for object-removal segments.
The training cohort contains 2,196 examples of 4--15 seconds, covering 1,969 object-preserving, 119 object-removal, and 108 object-introduction interactions.
Training excludes first- and last-frame image conditions.
For long evaluation streams, each product presentation is followed by removal, with the person-and-scene reference supplied as the removal segment's end-frame condition.

\paragraph{Metrics and Aggregation.}
CLIP and DINOv2 measure whole-frame reference similarity and temporal similarity between successive 1\,Hz samples~\citep{clip,dinov2}.
Full-prompt CLIP and ViCLIP assess text alignment; VABench measures realism and five-question case-level visual QA~\citep{vabench}.
At instruction changes, CLIP and DINO compare adjacent boundary frames; Luma is the absolute difference in their mean grayscale intensity (0--255).
Scores use equal case weights; paired bootstrapping resamples cases while retaining each complete stream and its matched model outputs.
We draw 100,000 paired resamples and report unadjusted percentile 95\% intervals.

\MainResultsTable
\SubgroupResultsTable
\subsection{Effect of the Specialization Axis}
\label{sec:ablation}
The same-backbone block of \cref{tab:main_results} compares all three assignments.
TaoMate-R1 achieves an aggregate DINO temporal similarity of 0.934, compared with 0.928 for Operation split and 0.914 for Shared.

\Cref{tab:subgroup_results} separates short interactions, long changing-instruction streams, and public reference compositions.
TaoMate-R1 obtains the highest mean CLIP and DINO temporal similarity in all three groups.
On short interactions, it also leads in DINO reference similarity, reaching 0.545 versus 0.531 for Operation split.
Its CLIP and DINO temporal similarities are 0.970 and 0.975, compared with 0.969 and 0.972 for Operation split, while Operation split has higher visual QA (0.900 versus 0.880).

On long streams, TaoMate-R1 reaches CLIP and DINO temporal similarities of 0.956 and 0.921, compared with 0.948 and 0.914 for Operation split.
The paired difference in DINO temporal similarity (TaoMate-R1 minus Operation split) is $+0.00730$ (95\% interval $[-0.00958,0.0227]$).
Operation split has higher DINO reference similarity (0.486 versus 0.478).
Its QA of 1.00 and TaoMate-R1's 0.980 correspond to 50 and 49 correct answers across five questions per case.
On public reference compositions, TaoMate-R1 leads all four reported metrics, with DINO reference similarity of 0.207, CLIP and DINO temporal similarities of 0.948 and 0.911, and visual QA of 0.983.

\paragraph{Interaction Settings and Shared Adaptation.}
Both two-adapter configurations improve DINO temporal similarity over Shared in every evaluated group.
At the same adapter budget, TaoMate-R1 further increases the temporal means across all three settings, supporting the assignment of adaptation capacity by opening and continuation.
The aggregate score combines twenty-two short or public cases with ten long streams under equal case weighting; the subgroup results show higher temporal means for TaoMate-R1 within each group.
Reference similarity and visual QA vary by setting, indicating that higher temporal similarity does not imply uniformly better reference matching or visual QA.

\Cref{fig:long_range_change} examines background appearance in fixed foliage and vase crops over a 119-second stream.
Views at 0, 60, and 119 seconds show the evolution of leaf and ceramic tones.
The overlays average per-pixel CIE Lab distances from each video's initial frame over the full stream at 1\,Hz.
TaoMate-R1 shows smaller appearance changes than Operation split in these regions, complementing the whole-frame temporal measurements.
\newcommand{\BackgroundDetailColumn}[2]{%
\begin{minipage}[t]{.322\linewidth}
\centering
\begin{minipage}[t]{.98\linewidth}
\centering
\makebox[.32\linewidth]{\scriptsize\sffamily 0\,s}\hfill
\makebox[.32\linewidth]{\scriptsize\sffamily 60\,s}\hfill
\makebox[.32\linewidth]{\scriptsize\sffamily 119\,s}\par
\vspace{1pt}
\includegraphics[width=.32\linewidth]{figures/assets/background_v26/#1_#2_t000.png}\hfill
\includegraphics[width=.32\linewidth]{figures/assets/background_v26/#1_#2_t060.png}\hfill
\includegraphics[width=.32\linewidth]{figures/assets/background_v26/#1_#2_t119.png}\par
\vspace{2pt}
\includegraphics[width=.90\linewidth]{figures/assets/background_v26/#1_#2_mean_overlay.png}
\end{minipage}
\end{minipage}%
}
\definecolor{BackgroundZero}{RGB}{248,248,249}
\definecolor{BackgroundThree}{RGB}{252,218,130}
\definecolor{BackgroundSix}{RGB}{247,157,64}
\definecolor{BackgroundNine}{RGB}{224,83,44}
\definecolor{BackgroundTwelve}{RGB}{155,24,45}
\pgfdeclarehorizontalshading{BackgroundScale}{100bp}{%
color(0bp)=(BackgroundZero);
color(25bp)=(BackgroundZero);
color(37.5bp)=(BackgroundThree);
color(50bp)=(BackgroundSix);
color(62.5bp)=(BackgroundNine);
color(75bp)=(BackgroundTwelve);
color(100bp)=(BackgroundTwelve)}
\begin{figure}[!t]
\centering
\begin{minipage}{.72\linewidth}
\makebox[.322\linewidth]{\scriptsize\sffamily\shortstack{Shared\\\strut}}\hfill
\makebox[.322\linewidth]{\scriptsize\sffamily\shortstack{Operation split\\\strut}}\hfill
\makebox[.322\linewidth]{\scriptsize\sffamily\shortstack{TaoMate-R1\\\strut}}\par
\vspace{2pt}
{\scriptsize\sffamily (a) Background foliage\par}
\vspace{1pt}
\BackgroundDetailColumn{foliage}{shared}\hfill
\BackgroundDetailColumn{foliage}{operation}\hfill
\BackgroundDetailColumn{foliage}{temporal}\par
\vspace{3pt}
{\scriptsize\sffamily (b) Ceramic vase\par}
\vspace{1pt}
\BackgroundDetailColumn{vase}{shared}\hfill
\BackgroundDetailColumn{vase}{operation}\hfill
\BackgroundDetailColumn{vase}{temporal}\par
\vspace{3pt}
{\scriptsize\sffamily Mean background change over 119\,s\quad $\overline{\Delta E}_{76}$\par}
\vspace{2pt}
\begin{tikzpicture}[x=\linewidth,y=1pt]
\path[use as bounding box] (0,-11) rectangle (1,3);
\shade[shading=BackgroundScale] (.12,0) rectangle (.88,3);
\foreach \v/\p in {0/.12,3/.31,6/.50,9/.69}{%
  \node[anchor=north,inner sep=0,font=\scriptsize\sffamily] at (\p,-2) {\v};}
\node[anchor=north,inner sep=0,font=\scriptsize\sffamily] at (.88,-2) {$12{+}$};
\end{tikzpicture}
\end{minipage}
\caption{\textbf{Background color and appearance over a long stream.}
TaoMate-R1 uses the Temporal split configuration.
Fixed foliage and vase crops from case 008, enlarged at 0, 60, and 119\,s.
Below each triplet, the overlay shows mean CIE Lab distance from the initial frame over 119 samples at 1\,Hz.
All methods use identical crop coordinates and a shared scale (saturated at 12; opacity 0.45).}
\label{fig:long_range_change}
\end{figure}
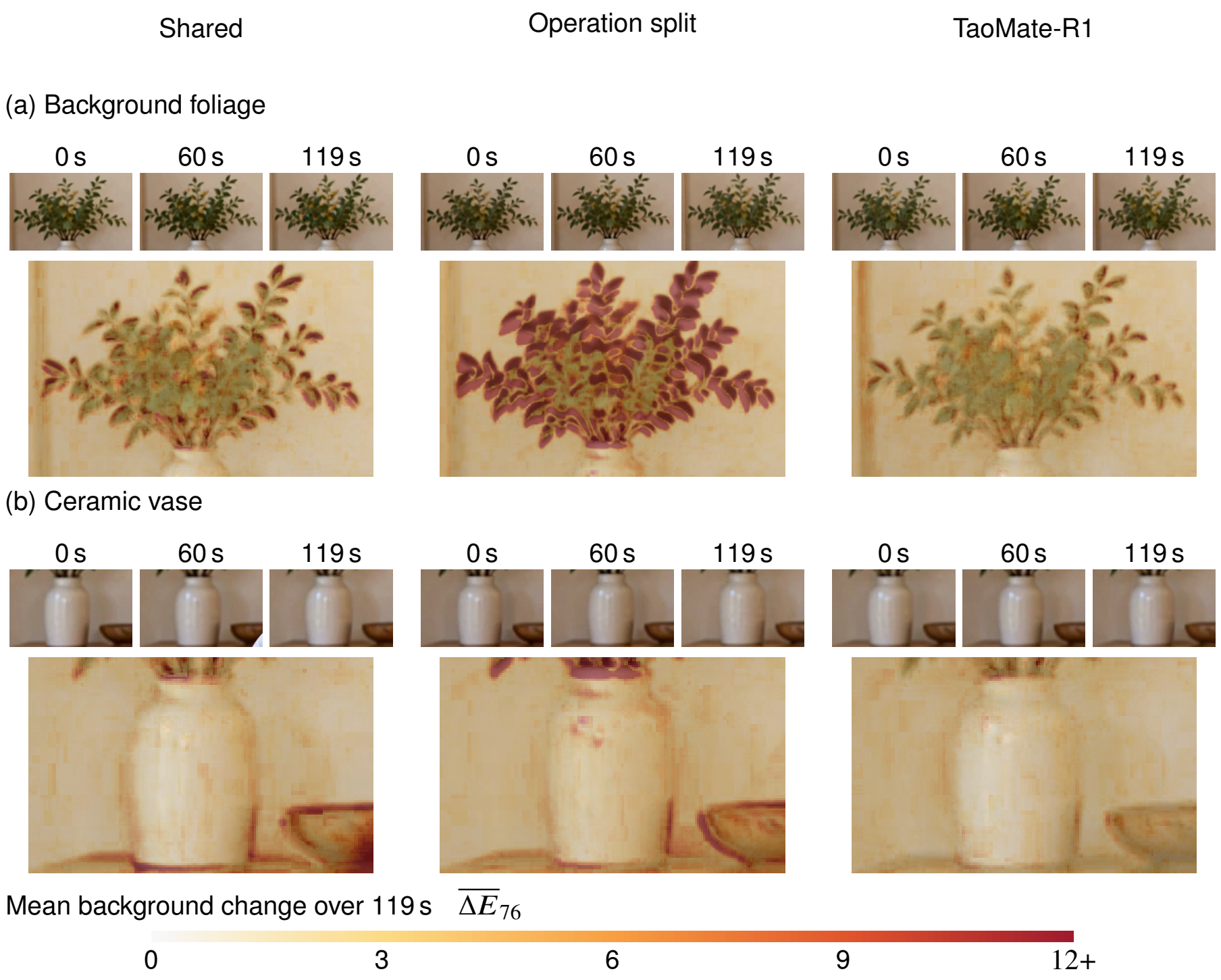

\paragraph{Interaction Dynamics.}
\Cref{fig:role_interactions} compares the three sharing schemes at matched timestamps within two changing-instruction streams.
In the bag sequence, all three configurations present the referenced item at 72 and 74\,s.
At 76\,s, TaoMate-R1 moves it toward the instructed screen-right side, Shared moves it screen-left, and Operation split still holds it centrally.
In the second sequence, Shared and TaoMate-R1 proceed from the notebook to the folded umbrella; Operation split remains empty-handed at 106 and 108\,s.
The TaoMate-R1 frames show evolving hand poses and object states within the retained scene.
Together with the background views, these sequences make both components of streaming interaction visible: carrying out a new action and preserving the scene in which it unfolds.
\newcommand{\RoleInteractionFrame}[5]{%
\node[anchor=north west,inner sep=0] at (#4,#5)
{\includegraphics[width=96pt]{figures/assets/interactions_v32/case#1_#2_t#3.jpg}};}
\newcommand{\RoleInteractionLabels}{%
\node[align=center,text width=44pt] at (85,-37) {Shared};
\node[align=center,text width=44pt] at (85,-96) {Operation\\split};
\node[align=center,text width=44pt] at (85,-155) {TaoMate-R1};}
\begin{figure*}[t]
\centering
\resizebox{\textwidth}{!}{%
\begin{tikzpicture}[x=1pt,y=1pt,font=\sffamily\fontsize{8}{9}\selectfont]
\path[use as bounding box] (0,-398) rectangle (505,25);
\node[anchor=west,font=\sffamily\bfseries\fontsize{9}{10}\selectfont] at (0,20)
{(a) Cosmetics-bag interaction};
\node at (30,0) {Inputs};
\foreach \t/\x in {70/157,72/257,74/357,76/457}
  \node at (\x,0) {\t\,s};
\RoleInteractionLabels
\node[inner sep=0] at (30,-45)
  {\includegraphics[width=60pt]{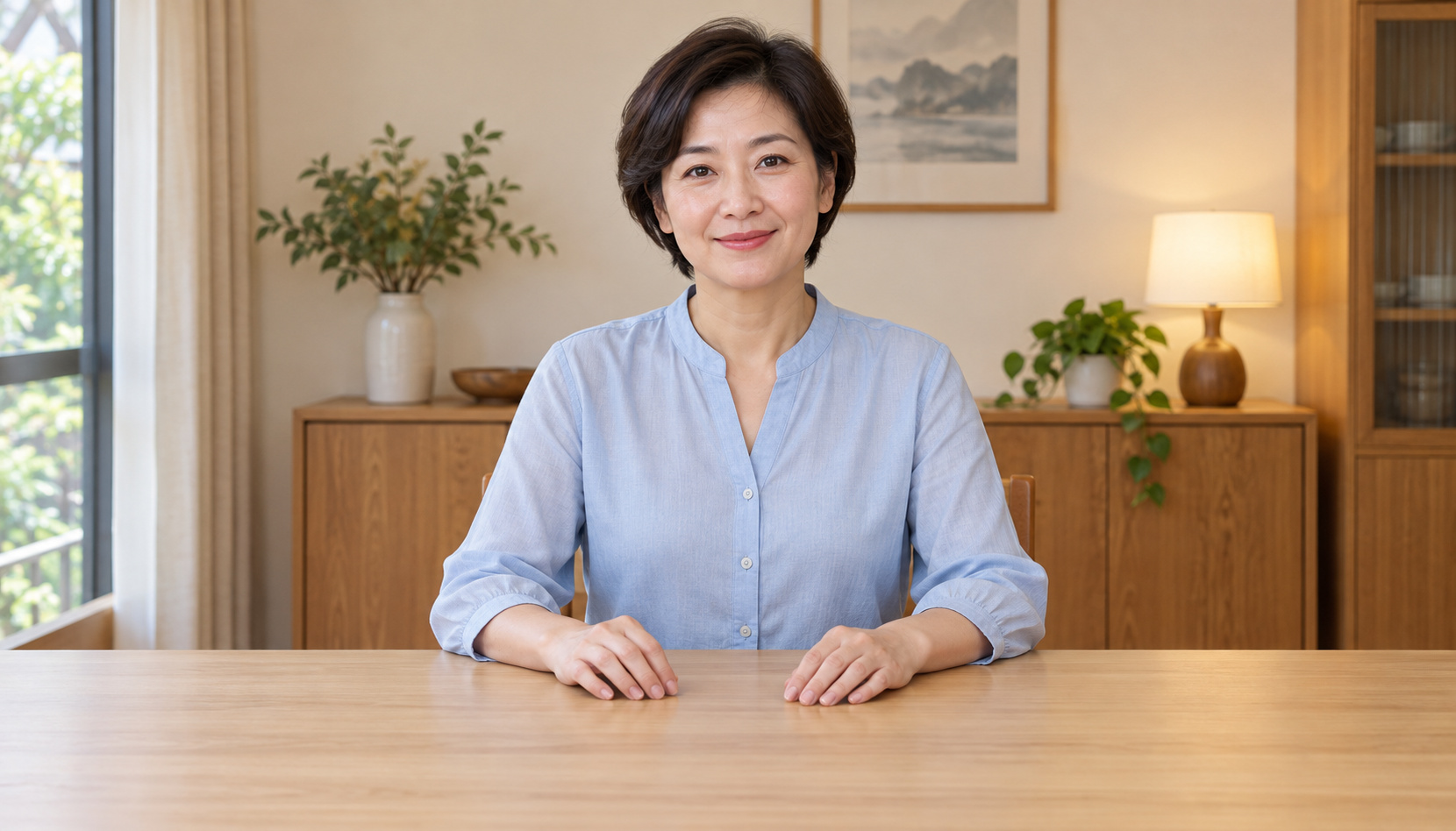}};
\node[inner sep=0] at (30,-109)
  {\includegraphics[width=57pt,trim={60 180 60 180},clip]{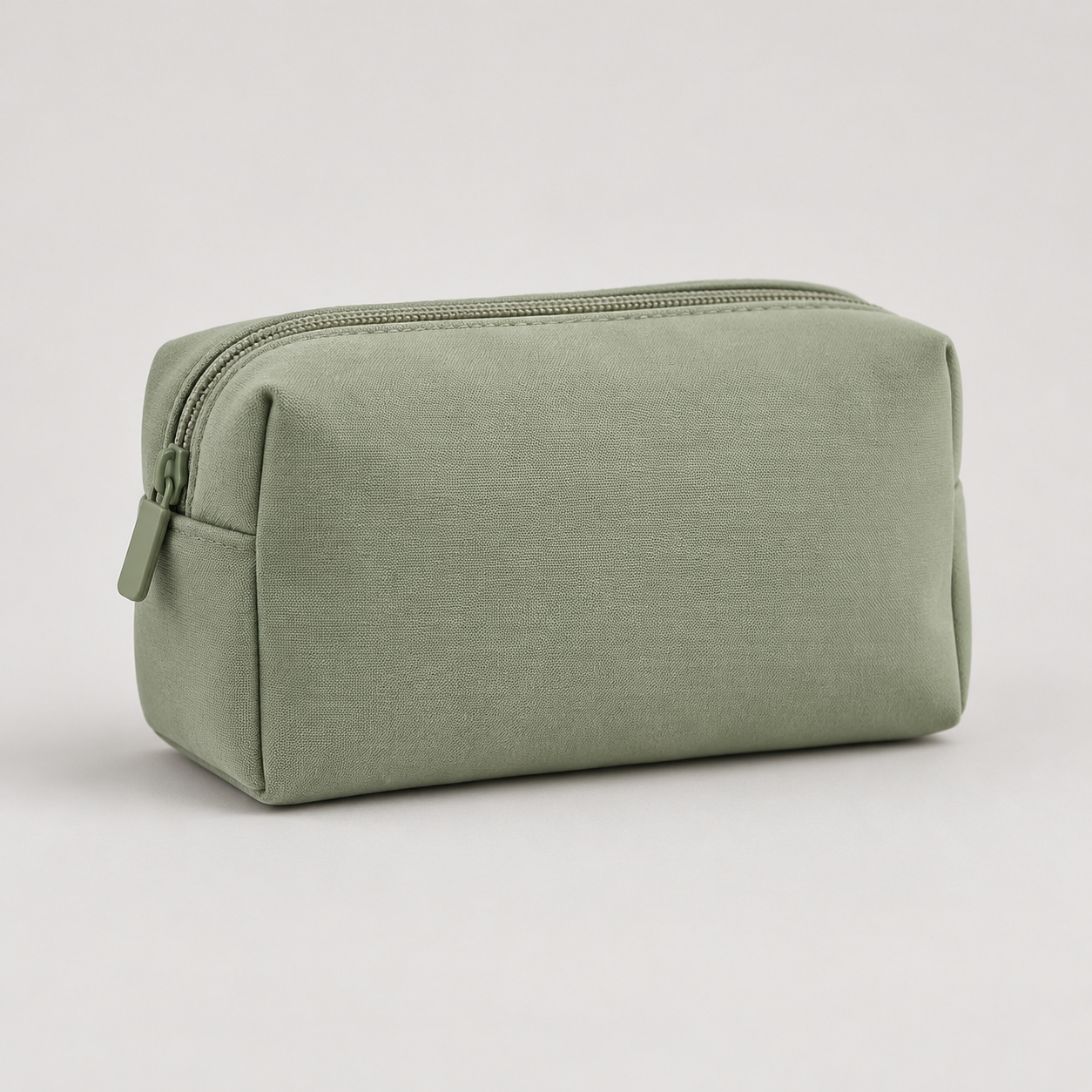}};
\node[align=center,text width=61pt,font=\sffamily\fontsize{7}{8}\selectfont] at (30,-158)
  {Present the bag\\then move it\\screen-right};
\foreach \method/\y in {shared/-10,operation/-69,temporal/-128}{
  \RoleInteractionFrame{008}{\method}{070}{109}{\y}
  \RoleInteractionFrame{008}{\method}{072}{209}{\y}
  \RoleInteractionFrame{008}{\method}{074}{309}{\y}
  \RoleInteractionFrame{008}{\method}{076}{409}{\y}}
\draw[black!20,line width=.35pt] (0,-196)--(505,-196);
\begin{scope}[yshift=-215pt]
\node[anchor=west,font=\sffamily\bfseries\fontsize{9}{10}\selectfont] at (0,10)
{(b) Notebook-to-umbrella transition};
\node at (30,-3) {Inputs};
\foreach \t/\x in {98/157,102/257,106/357,108/457}
  \node at (\x,-3) {\t\,s};
\RoleInteractionLabels
\node[inner sep=0] at (30,-39)
  {\includegraphics[width=60pt]{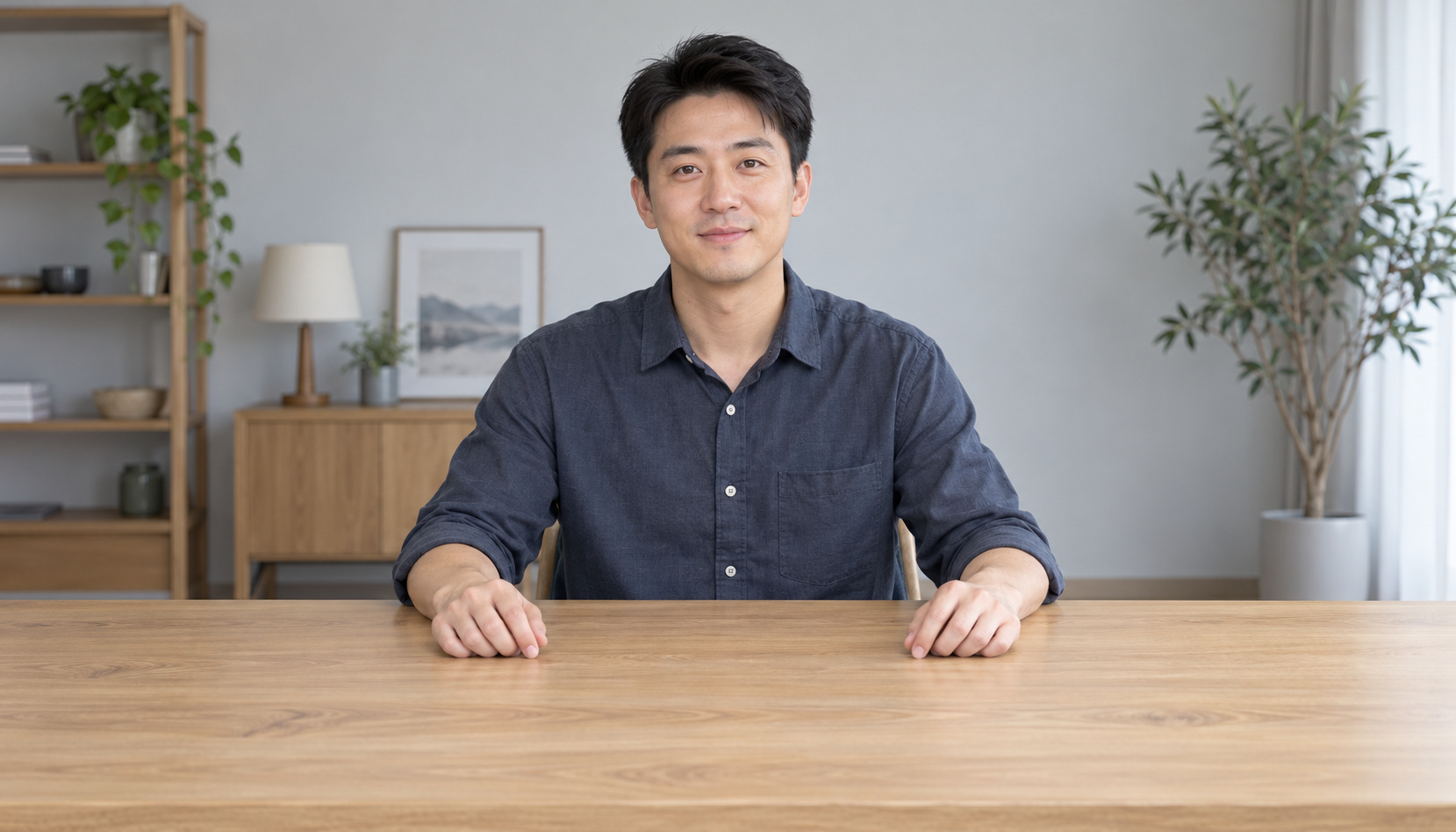}};
\node[inner sep=0] at (17,-99)
  {\includegraphics[width=28pt,trim={270 110 280 110},clip]{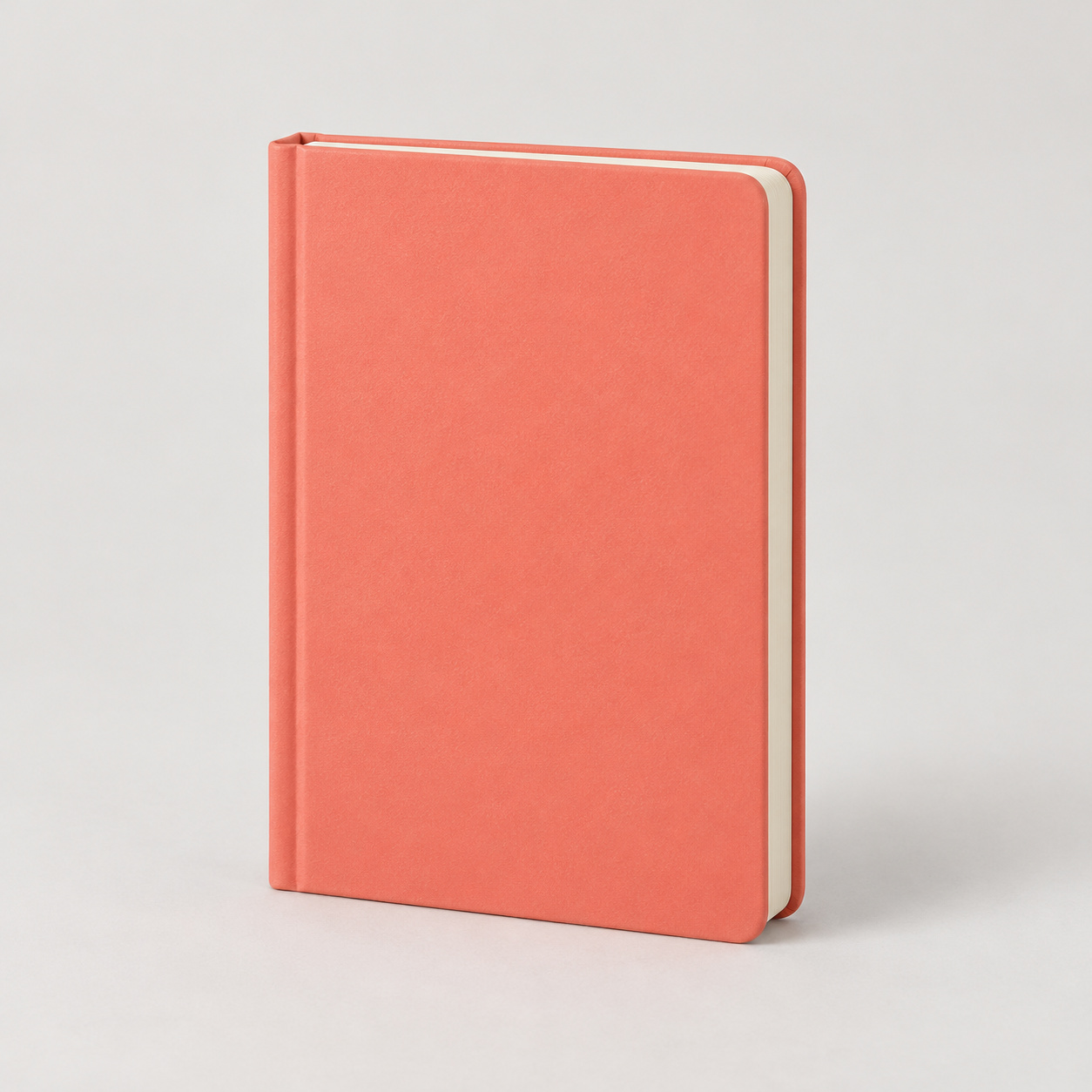}};
\node[inner sep=0] at (48,-99)
  {\includegraphics[width=23pt,trim={430 100 390 80},clip]{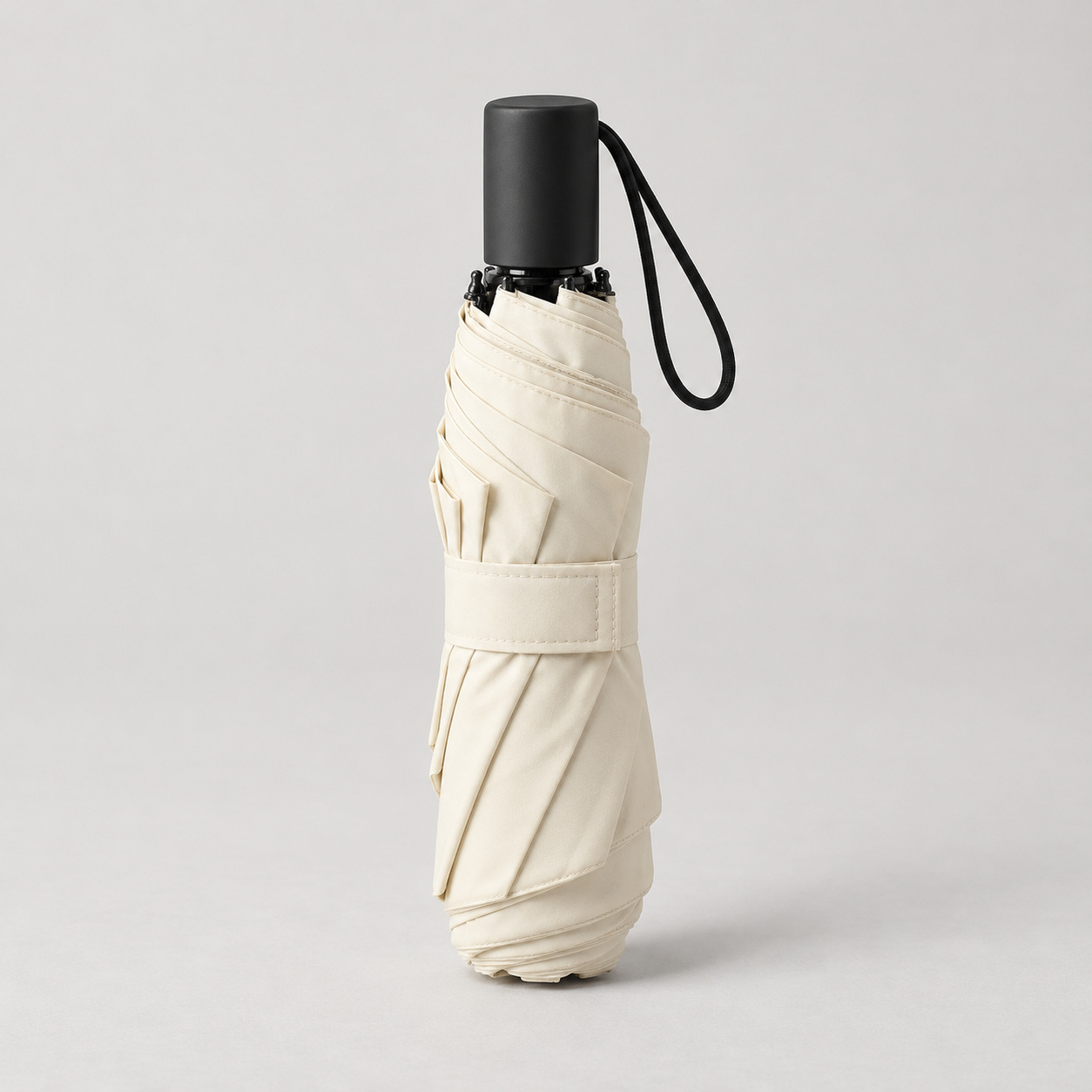}};
\node[align=center,text width=61pt,font=\sffamily\fontsize{7}{8}\selectfont] at (30,-158)
  {Present notebook\\then umbrella};
\foreach \method/\y in {shared/-10,operation/-69,temporal/-128}{
  \RoleInteractionFrame{006}{\method}{098}{109}{\y}
  \RoleInteractionFrame{006}{\method}{102}{209}{\y}
  \RoleInteractionFrame{006}{\method}{106}{309}{\y}
  \RoleInteractionFrame{006}{\method}{108}{409}{\y}}
\end{scope}
\end{tikzpicture}}
\caption{\textbf{Interaction sequences under three parameter-sharing schemes.}
TaoMate-R1 uses Temporal split.
Original references are at left; each column shows the same timestamp across all three H3 configurations.
(a) Bag presentation and the start of a screen-right set-aside action (case 008).
(b) Successive notebook and folded-umbrella presentations (case 006).
TaoMate-R1 changes the held object and interaction state while retaining the presenter and room; Operation split remains empty-handed in the sampled umbrella segment.}
\label{fig:role_interactions}
\end{figure*}
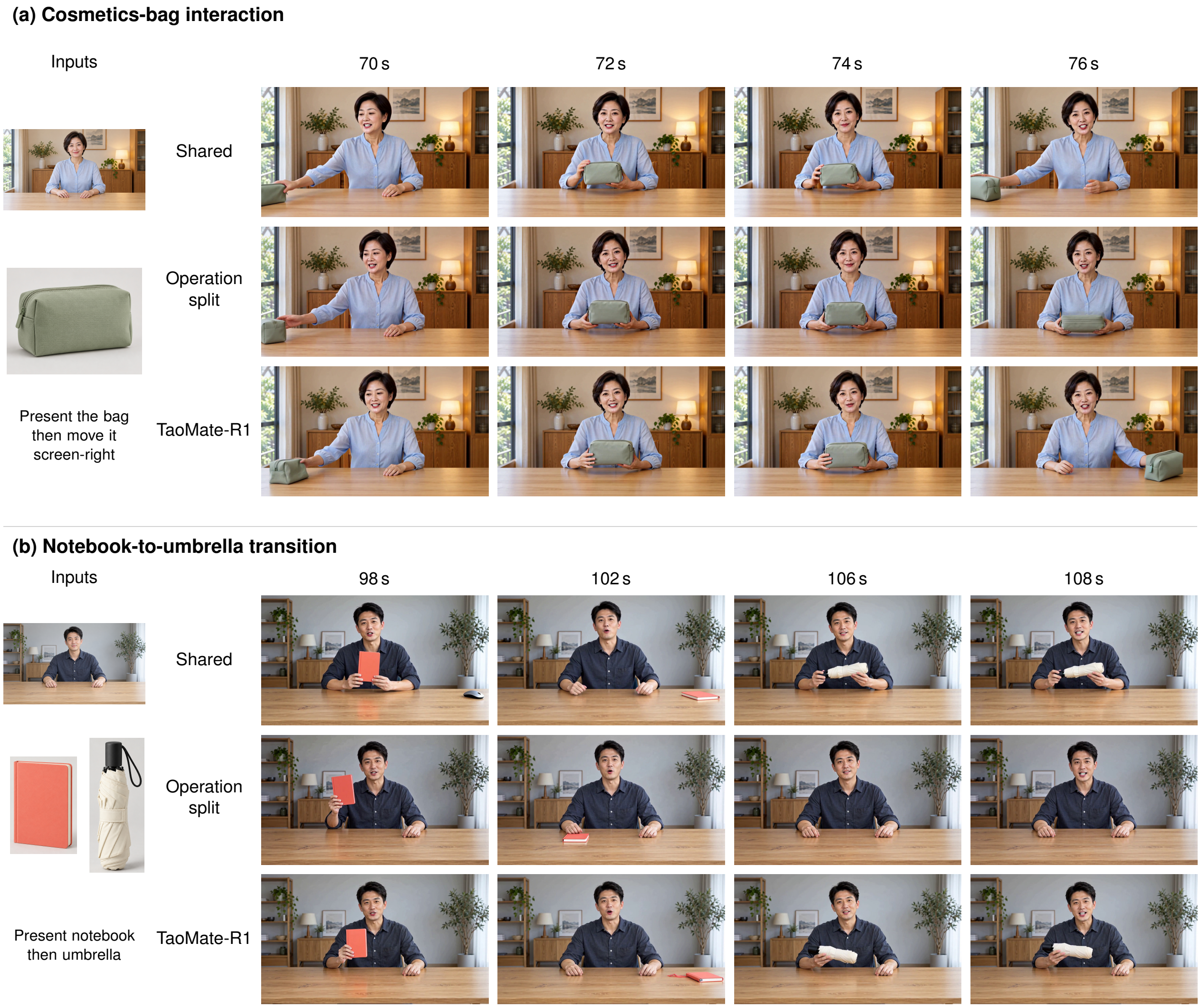

\subsection{Continuity across Instruction Changes}
\BoundaryResultsTable
\Cref{tab:boundary_results} compares adjacent frames at the same instruction boundaries in the ten long streams.
TaoMate-R1 obtains CLIP and DINO boundary similarities of 0.991 and 0.990, compared with 0.979 and 0.973 for Operation split.
Its mean luminance jump is 0.613 versus 1.96, a reduction of 1.35 gray levels, or 68.7\%.
Shared obtains boundary similarities of 0.979 and 0.978 and a mean luminance jump of 1.86.

The case-paired analysis gives a difference (TaoMate-R1 minus Operation split) of $+0.0177$ in DINO boundary similarity (95\% interval $[0.00820,0.0288]$) and $-1.35$ in luminance jump ($[-1.58,-1.12]$).
Luminance jumps decrease in all ten streams, and CLIP and DINO boundary similarities each increase in nine streams.
These transition-level gains are more consistent across cases than the sequence-level DINO temporal difference.

\paragraph{Continuity under Changing Conditions.}
An instruction change introduces a new interaction goal while the preceding scene remains the starting state for continuation.
Boundary measurements evaluate the immediate transition from the last frame under one instruction to the first under the next.
Sequence-level temporal scores also include the intervening within-instruction evolution.
Together, they show higher temporal-similarity means over the full streams, with the clearest paired evidence at instruction changes.
The background views in \cref{fig:long_range_change} extend this analysis to retained scene appearance over the complete stream.
These observations support temporal specialization for maintaining scene continuity as interaction goals change.

\subsection{Generation Quality}
\paragraph{Reference-Conditioned Systems.}
The external-system block of \cref{tab:main_results} compares MiniMax-H3~\citep{minimaxh3}, VACE~\citep{vace}, SkyReels-V3~\citep{skyreelsv3}, and Phantom~\citep{phantom}.
For the long cases, these systems generate each instruction segment independently and concatenate the resulting clips; our autoregressive variants retain history across instruction changes.
TaoMate-R1 leads in mean DINO reference similarity, both temporal similarity metrics, realism, and visual QA.
VACE leads in CLIP reference and full-prompt CLIP scores, and MiniMax-H3 in ViCLIP.

\paragraph{Qualitative Comparisons.}
\Cref{fig:qualitative_v10} shows four reference-conditioned interactions.
In the gallery scene, MiniMax-H3 and TaoMate-R1 depict both cartoon characters interacting with a guitar.
VACE renders live-action participants, while SkyReels-V3 and Phantom depict Santa as a small held figure.
In the scarf scene, our output includes an additional guitar.
The continuous-interaction rows show cosmetics-bag and closed-umbrella presentation within the established scene.
At 72\,s, our stream depicts the sage-green bag alongside the presenter in the established room.
\newlength{\QualVThirteenWidth}
\newlength{\QualVThirteenHeight}
\newlength{\QualVThirteenRefWidth}
\setlength{\QualVThirteenWidth}{.160\textwidth}
\setlength{\QualVThirteenHeight}{.090\textwidth}
\setlength{\QualVThirteenRefWidth}{.170\textwidth}
\newcommand{\QualVThirteenFrame}[2]{%
\includegraphics[width=\QualVThirteenWidth,trim={#2},clip]{figures/assets/results_v9/#1}%
}
\newcommand{\QualVEighteenFrame}[2]{%
\includegraphics[width=\QualVThirteenWidth,trim={#2},clip]{figures/assets/results_v18/#1}%
}
\newcommand{\QualVTwentyFrame}[2]{%
\includegraphics[width=\QualVThirteenWidth,trim={#2},clip]{figures/assets/results_v20/#1}%
}
\newcommand{\QualVThirteenPublicRefs}[6]{%
\begin{tikzpicture}[baseline=(current bounding box.south)]
\path[use as bounding box] (0,0) rectangle (\QualVThirteenRefWidth,\QualVThirteenHeight);
\node[inner sep=0,anchor=west] at (0,.5\QualVThirteenHeight)
  {\includegraphics[width=.051\textwidth,trim={#2},clip]{figures/assets/results_v9/#1}};
\node[inner sep=0,anchor=west] at (.054\textwidth,.5\QualVThirteenHeight)
  {\includegraphics[width=.051\textwidth,trim={#4},clip]{figures/assets/results_v9/#3}};
\node[inner sep=0,anchor=west] at (.110\textwidth,.5\QualVThirteenHeight)
  {\includegraphics[width=.060\textwidth,height=.084\textwidth,keepaspectratio,
    trim={#6},clip]{figures/assets/results_v9/#5}};
\end{tikzpicture}%
}
\newcommand{\QualVThirteenSceneRef}[3]{%
\begin{tikzpicture}[baseline=(current bounding box.south)]
\path[use as bounding box] (0,0) rectangle (\QualVThirteenRefWidth,\QualVThirteenHeight);
\node[inner sep=0,anchor=west] at (0,.5\QualVThirteenHeight)
  {\includegraphics[width=.104\textwidth]{figures/assets/results_v9/#1}};
\node[inner sep=0,anchor=west] at (.109\textwidth,.5\QualVThirteenHeight)
  {\includegraphics[width=.061\textwidth,height=.086\textwidth,keepaspectratio,
    trim={#3},clip]{#2}};
\end{tikzpicture}%
}
\begin{figure*}[!htbp]
\centering
\begingroup
\setlength{\tabcolsep}{1.2pt}
\small
\begin{tabular}{@{}cccccc@{}}
\textbf{Inputs} & \textbf{MiniMax-H3} & \textbf{VACE} & \textbf{SkyReels-V3} & \textbf{Phantom} & \textbf{TaoMate-R1}\\[2pt]
\QualVThirteenPublicRefs{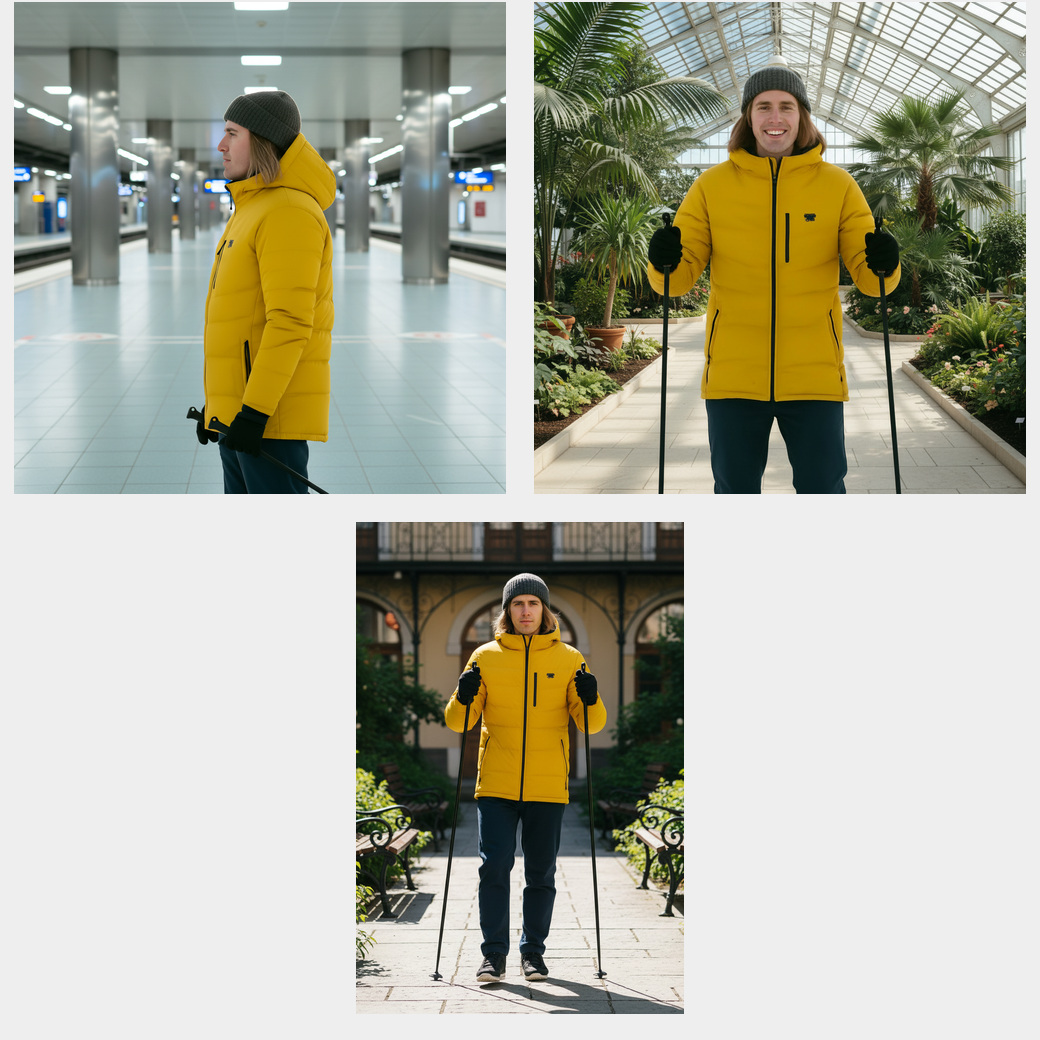}{534 546 12 0}{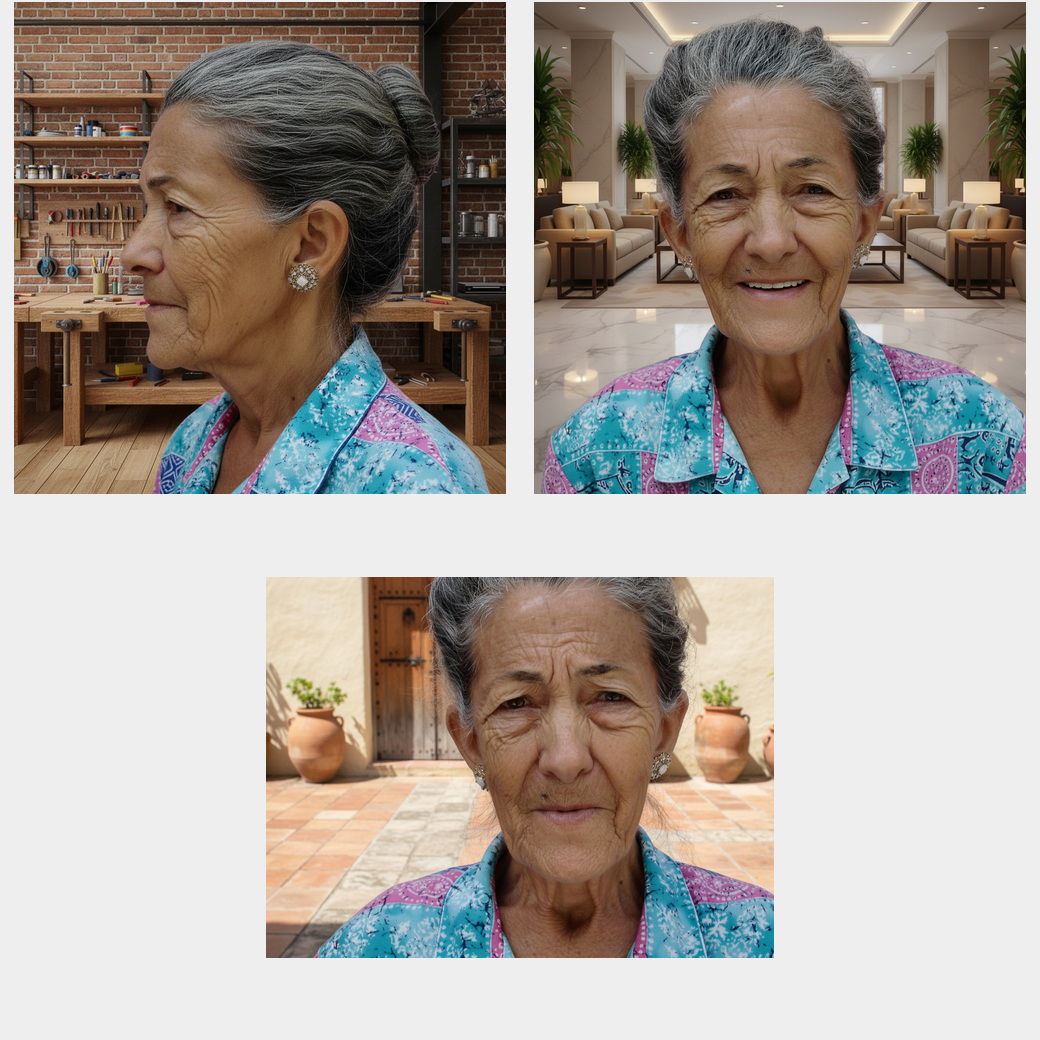}{534 546 14 2}{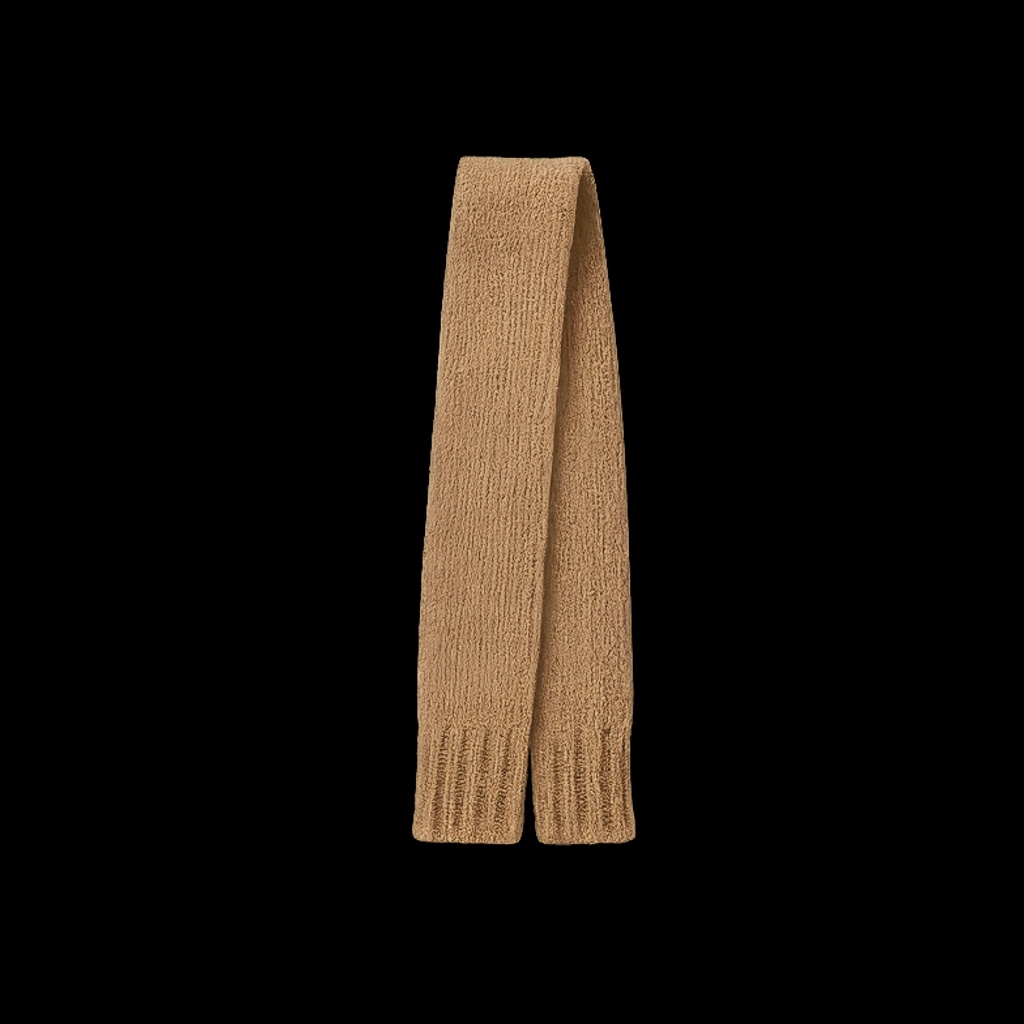}{380 140 360 140} &
\QualVThirteenFrame{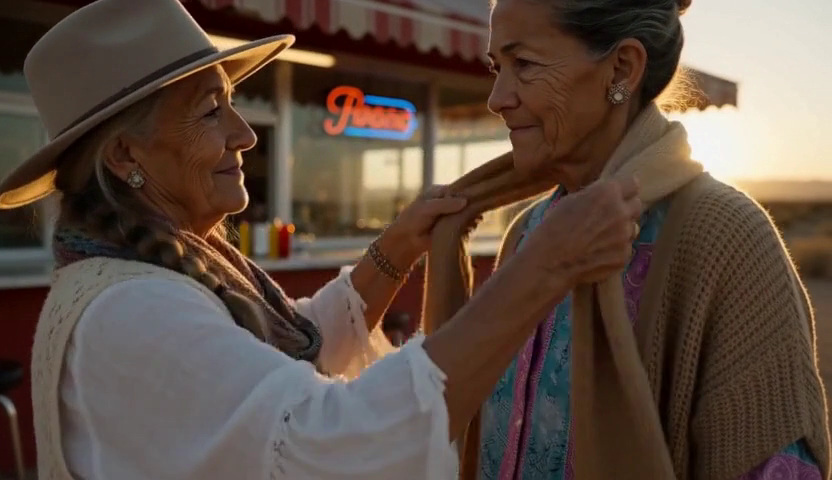}{0 6 0 6} &
\QualVThirteenFrame{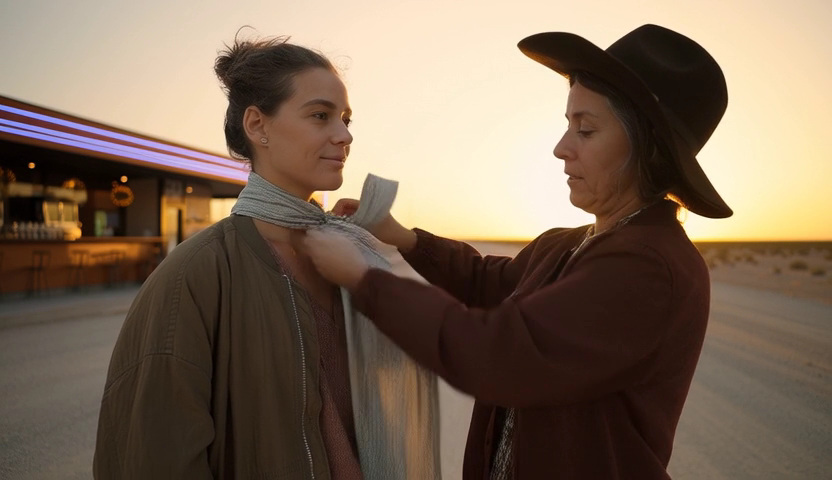}{0 6 0 6} &
\QualVThirteenFrame{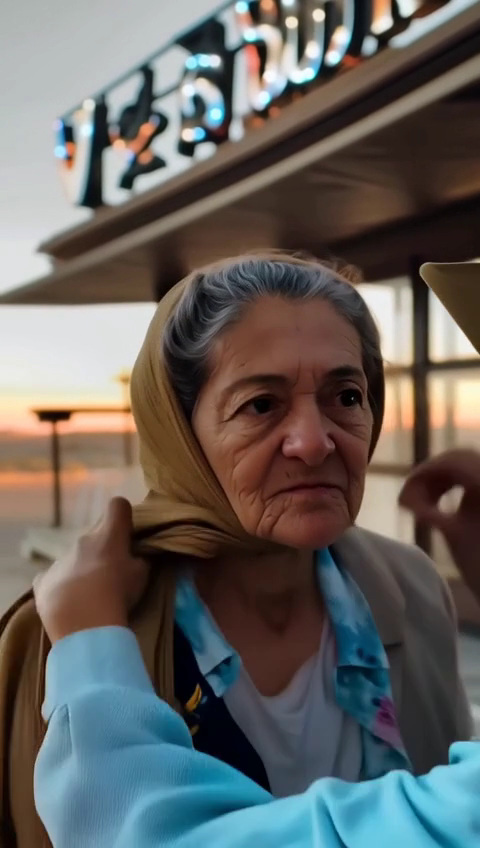}{0 218 0 360} &
\QualVThirteenFrame{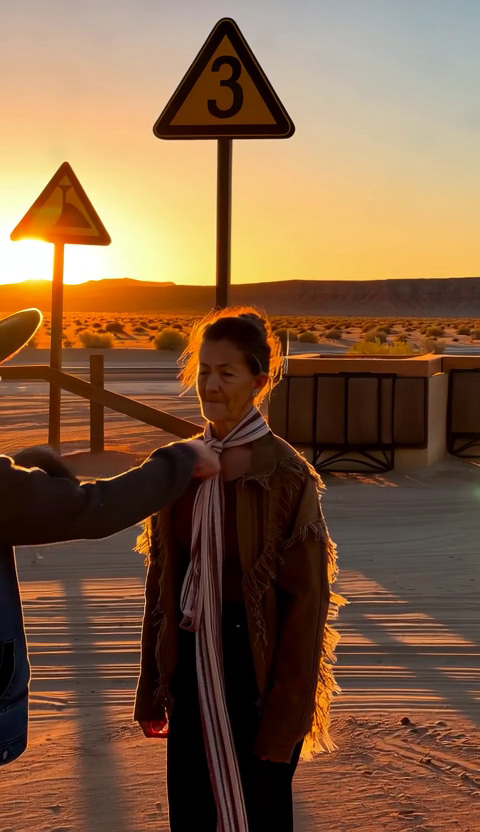}{0 262 0 300} &
\QualVThirteenFrame{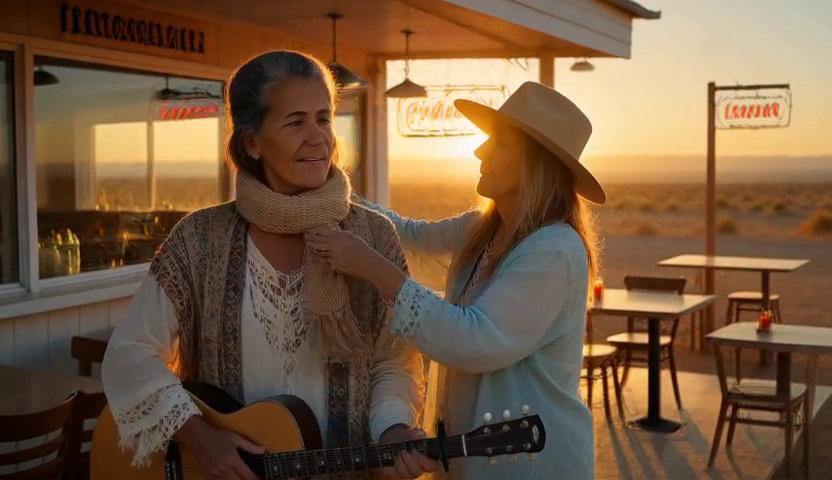}{0 6 0 6}\\[2pt]
\QualVThirteenPublicRefs{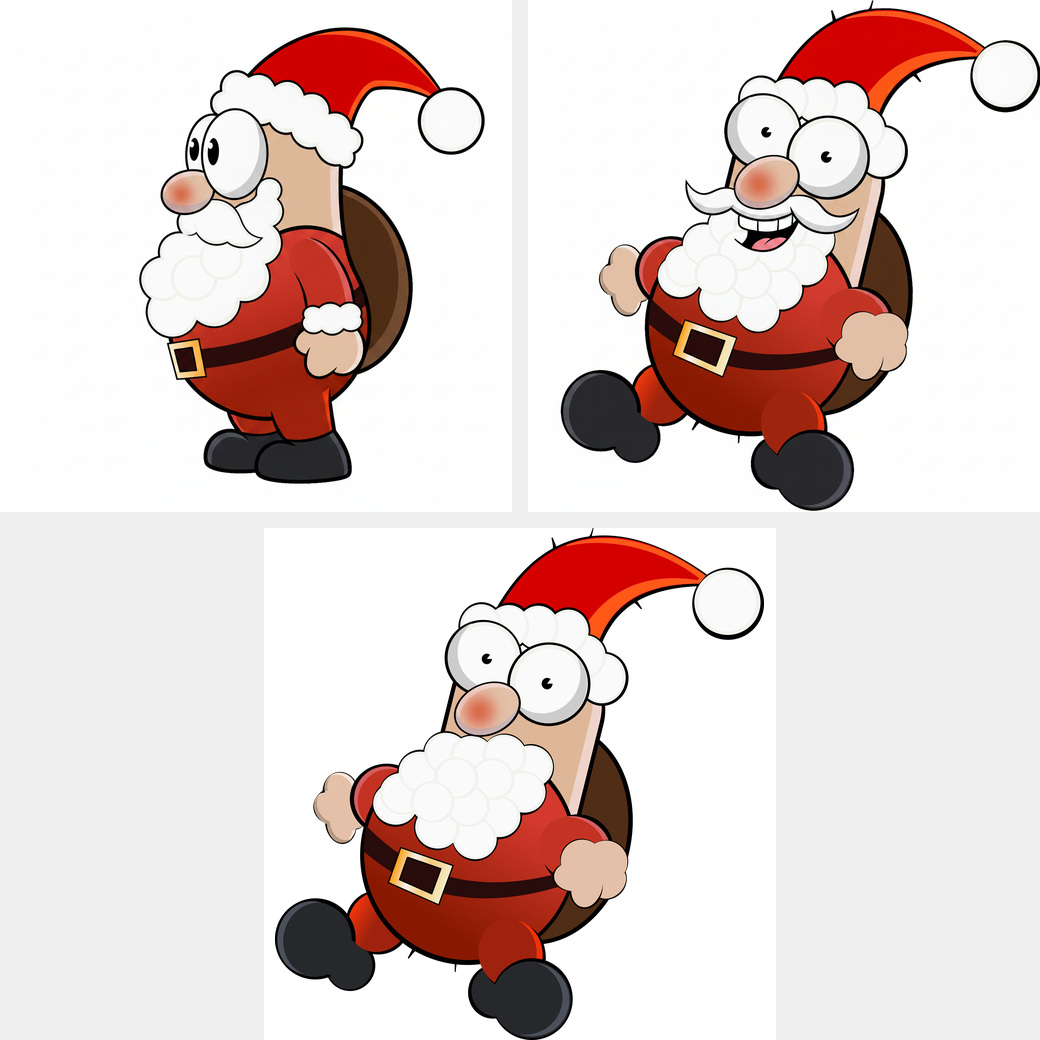}{0 528 528 0}{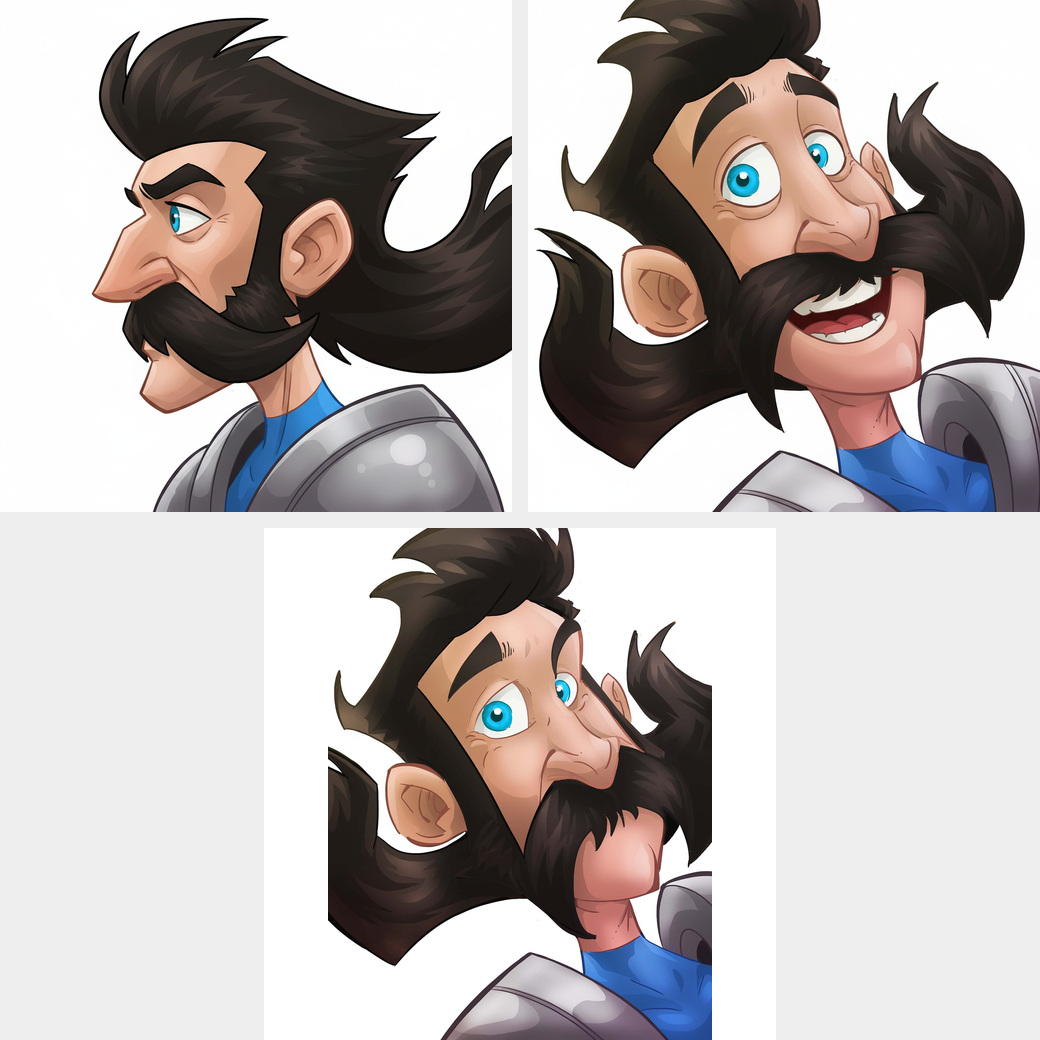}{0 528 528 0}{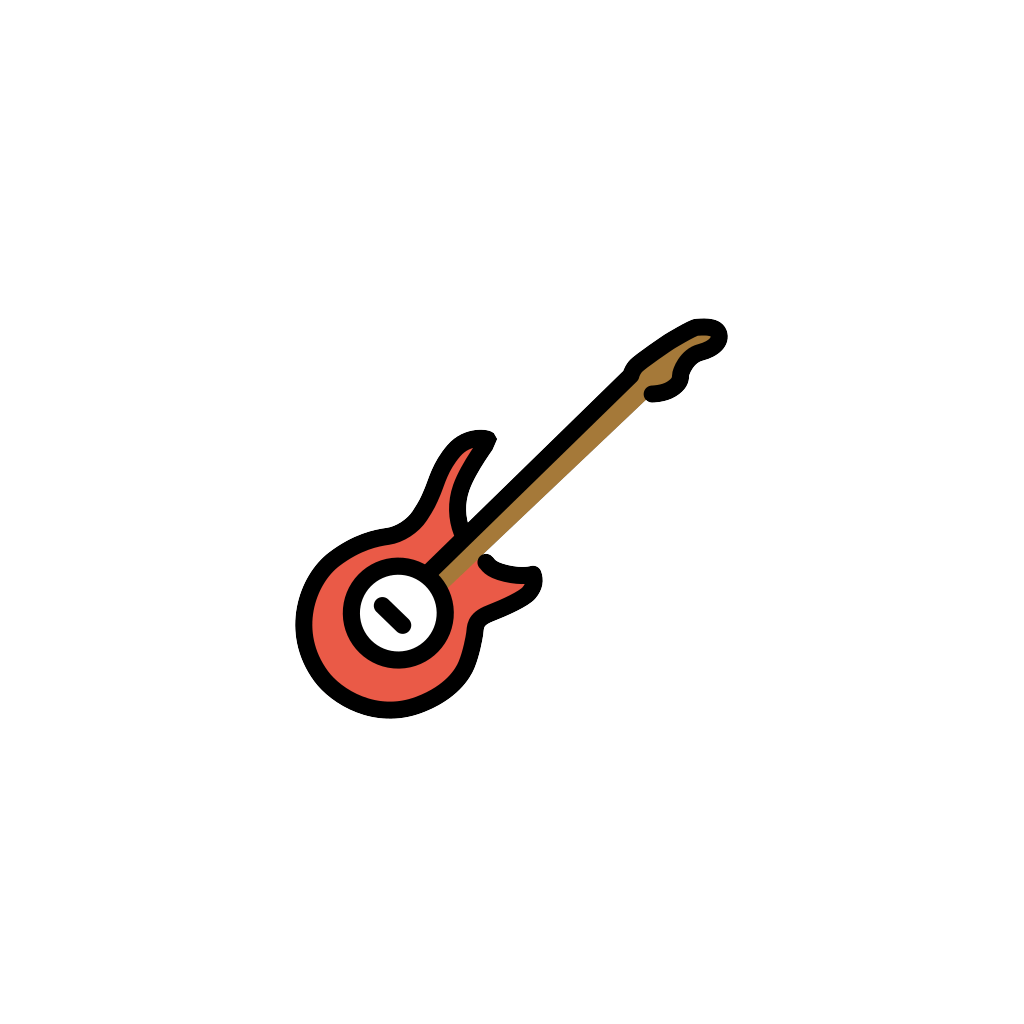}{300 280 300 320} &
\QualVEighteenFrame{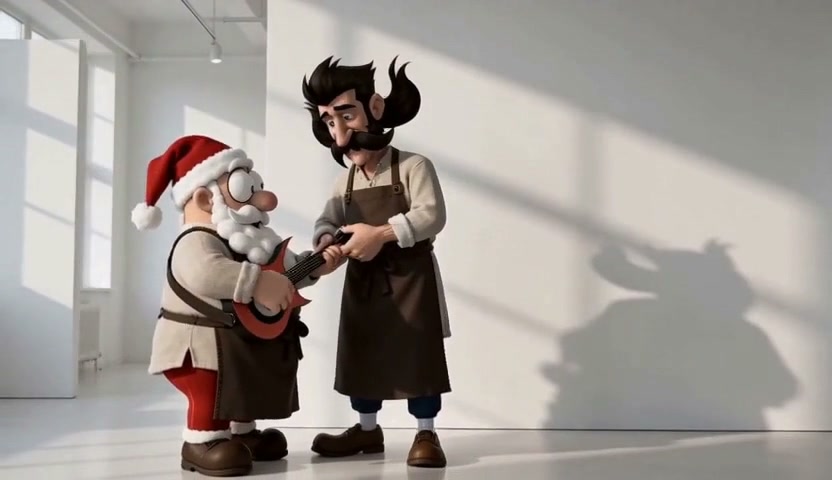}{60 115 212 50} &
\QualVEighteenFrame{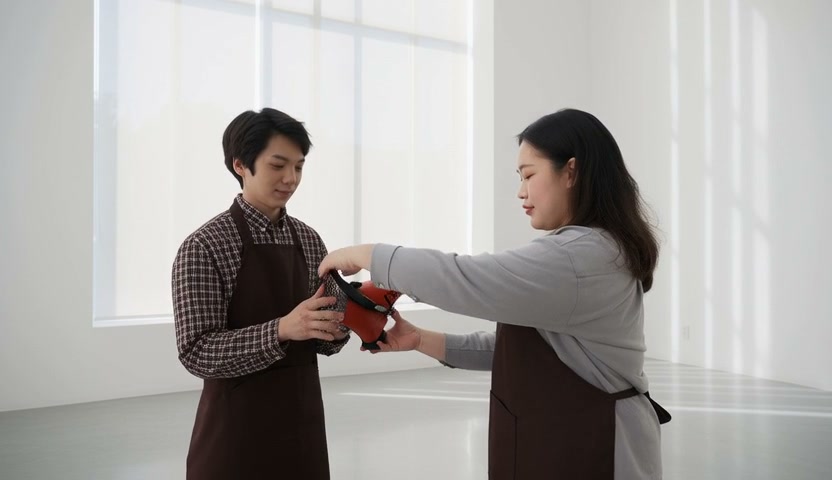}{0 6 0 6} &
\QualVEighteenFrame{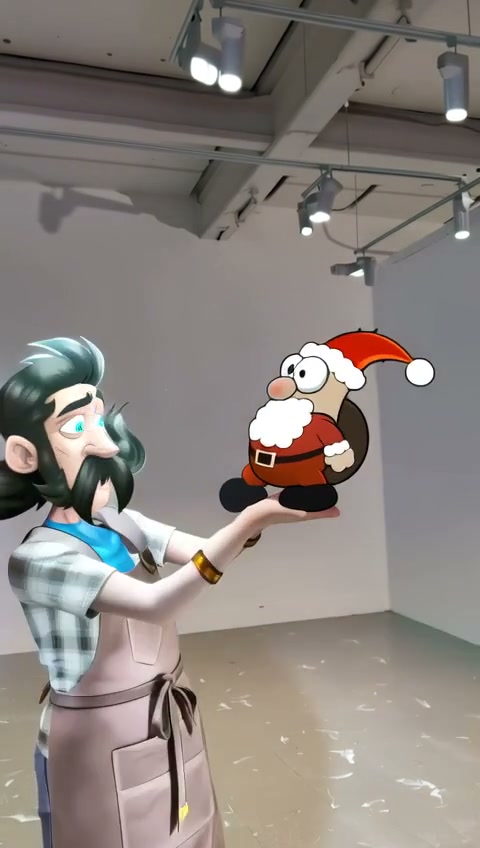}{0 263 0 315} &
\QualVEighteenFrame{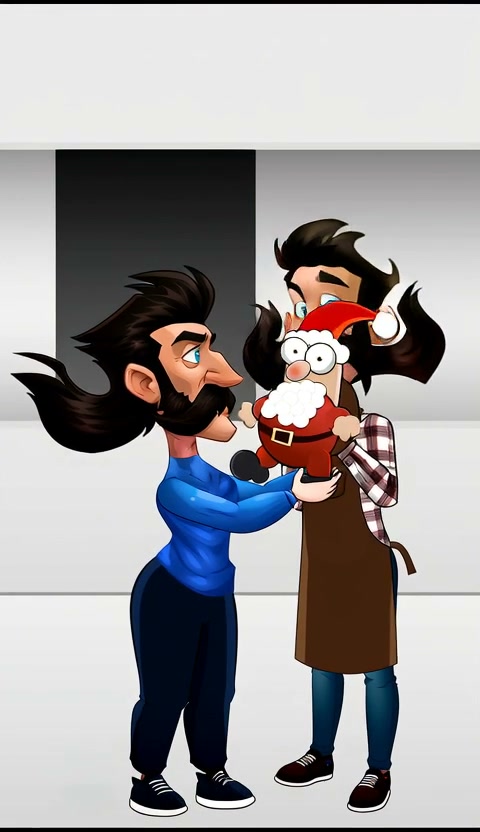}{0 332 0 230} &
\QualVEighteenFrame{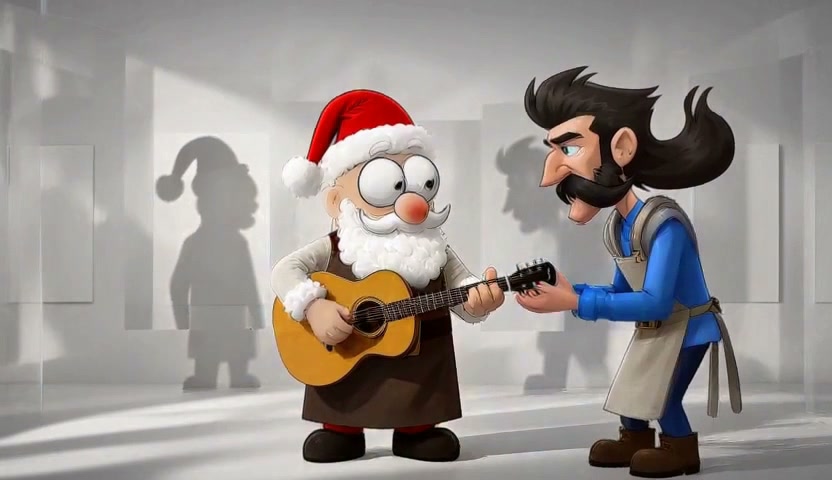}{0 6 0 6}\\[2pt]
\QualVThirteenSceneRef{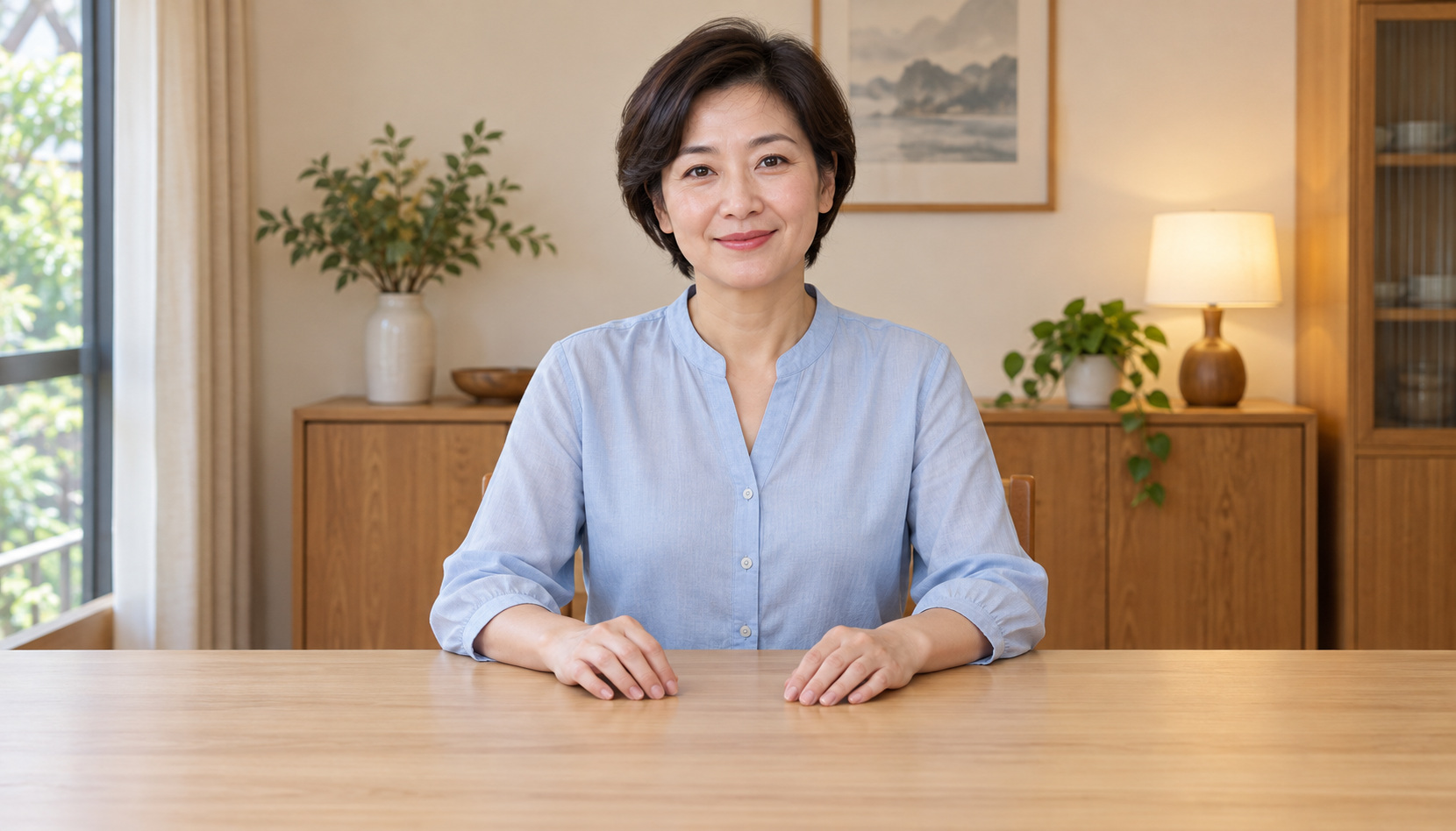}{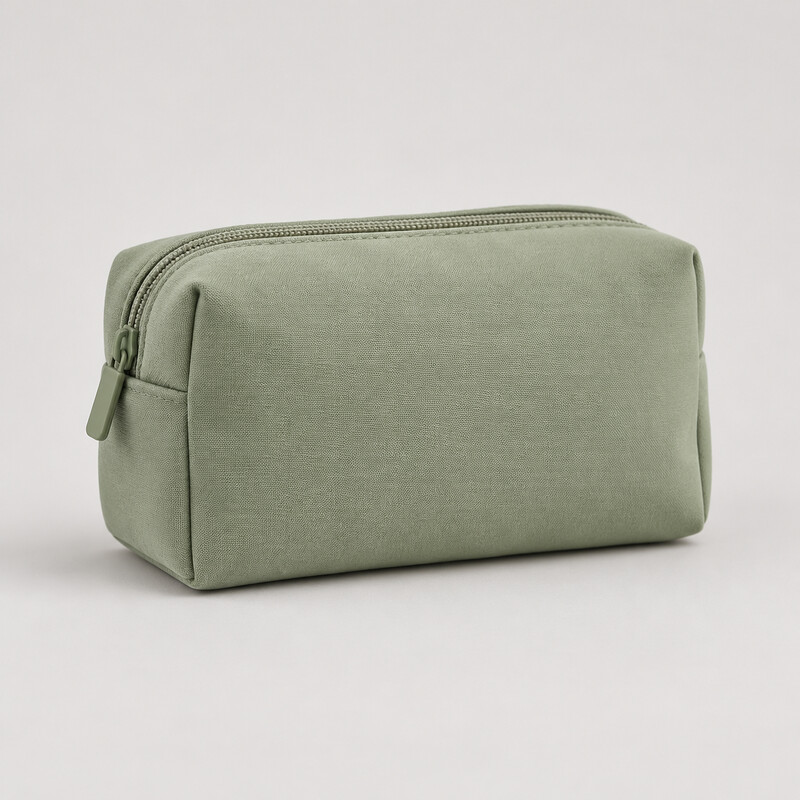}{60 180 60 180} &
\QualVTwentyFrame{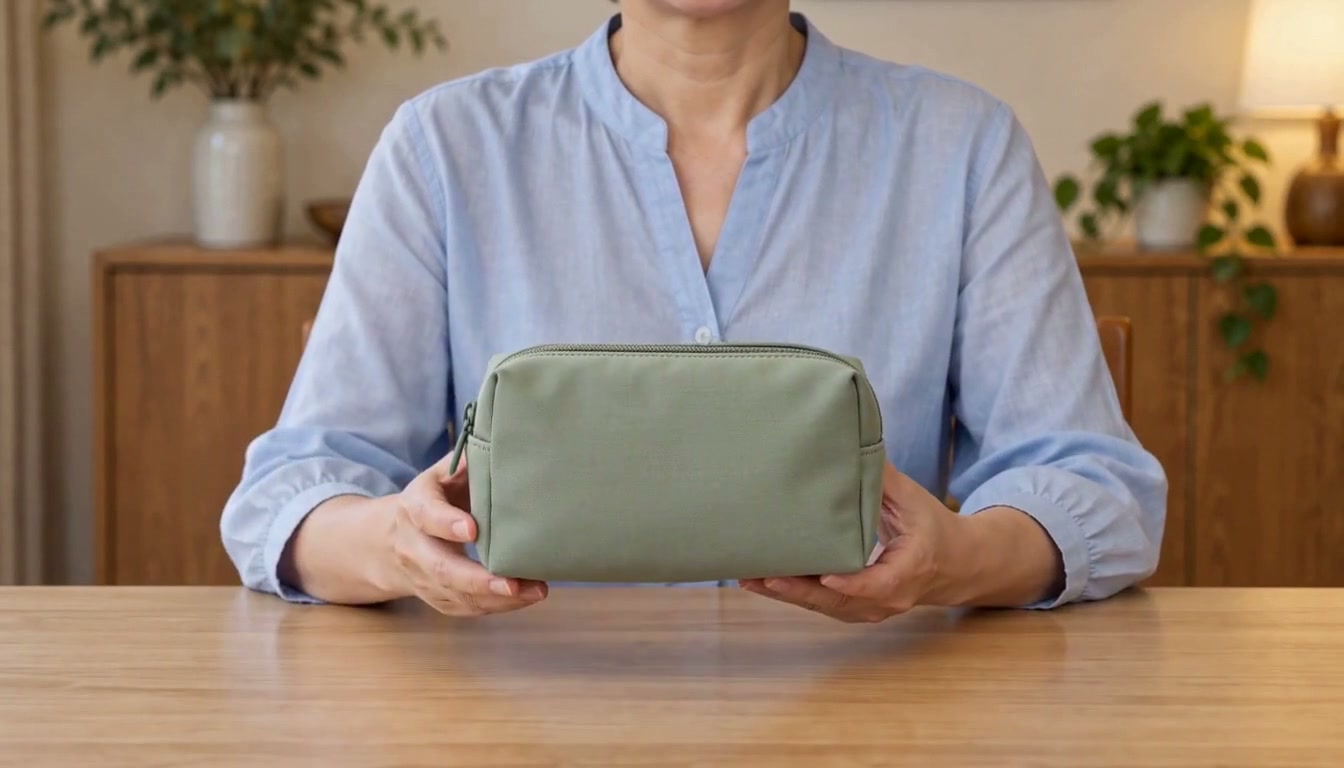}{0 6 0 6} &
\QualVTwentyFrame{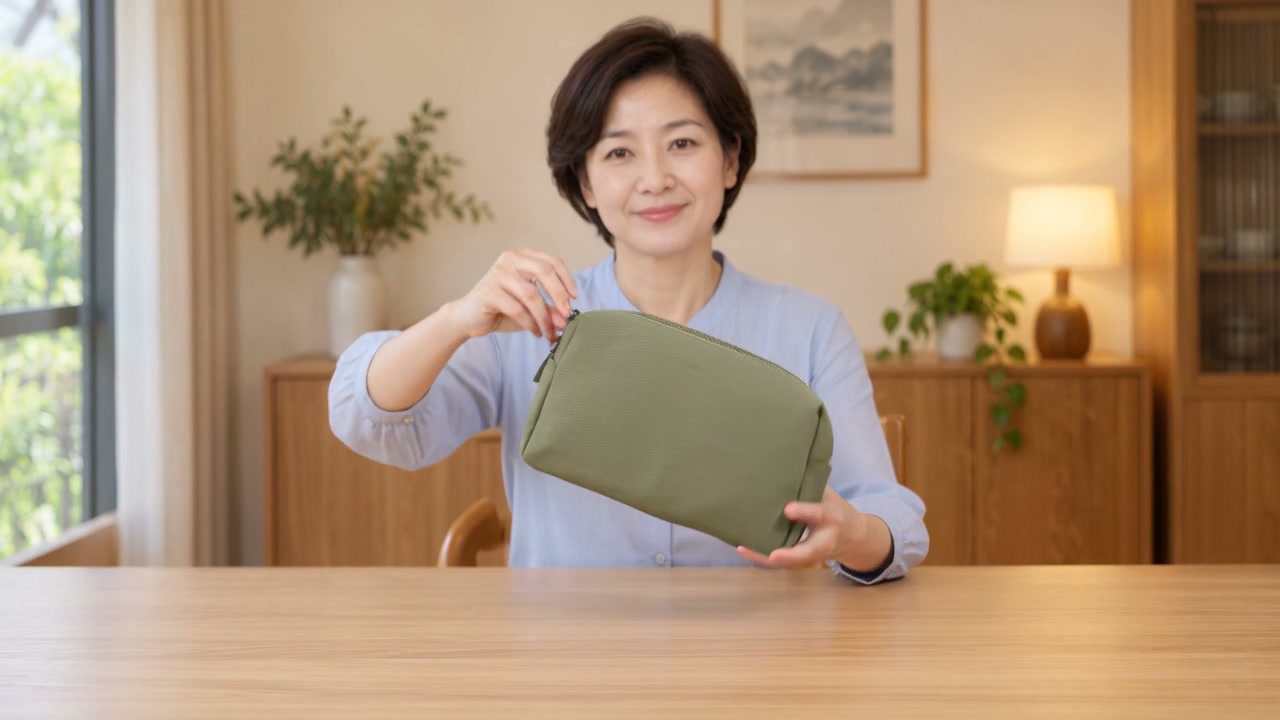}{0 0 0 0} &
\QualVTwentyFrame{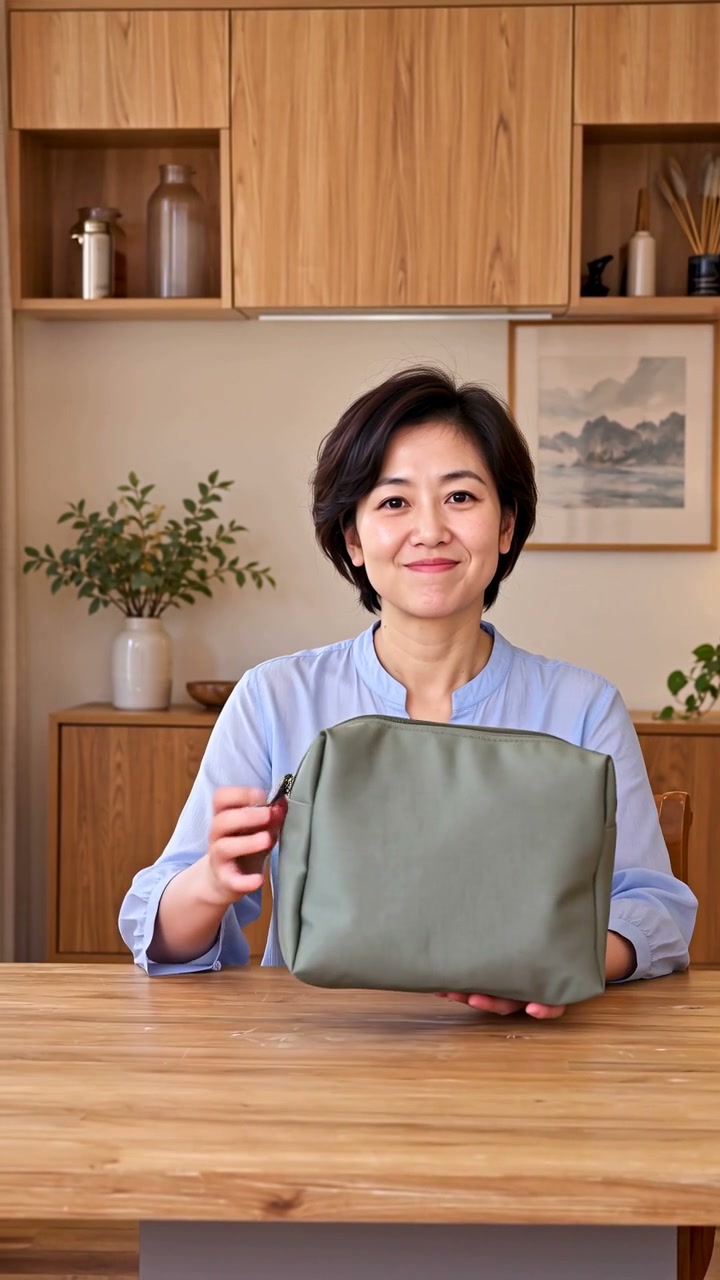}{0 240 0 635} &
\QualVTwentyFrame{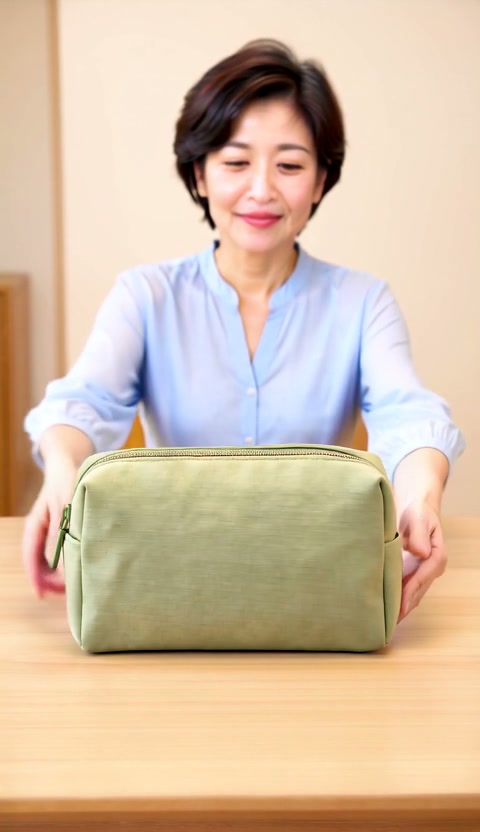}{0 160 0 402} &
\QualVTwentyFrame{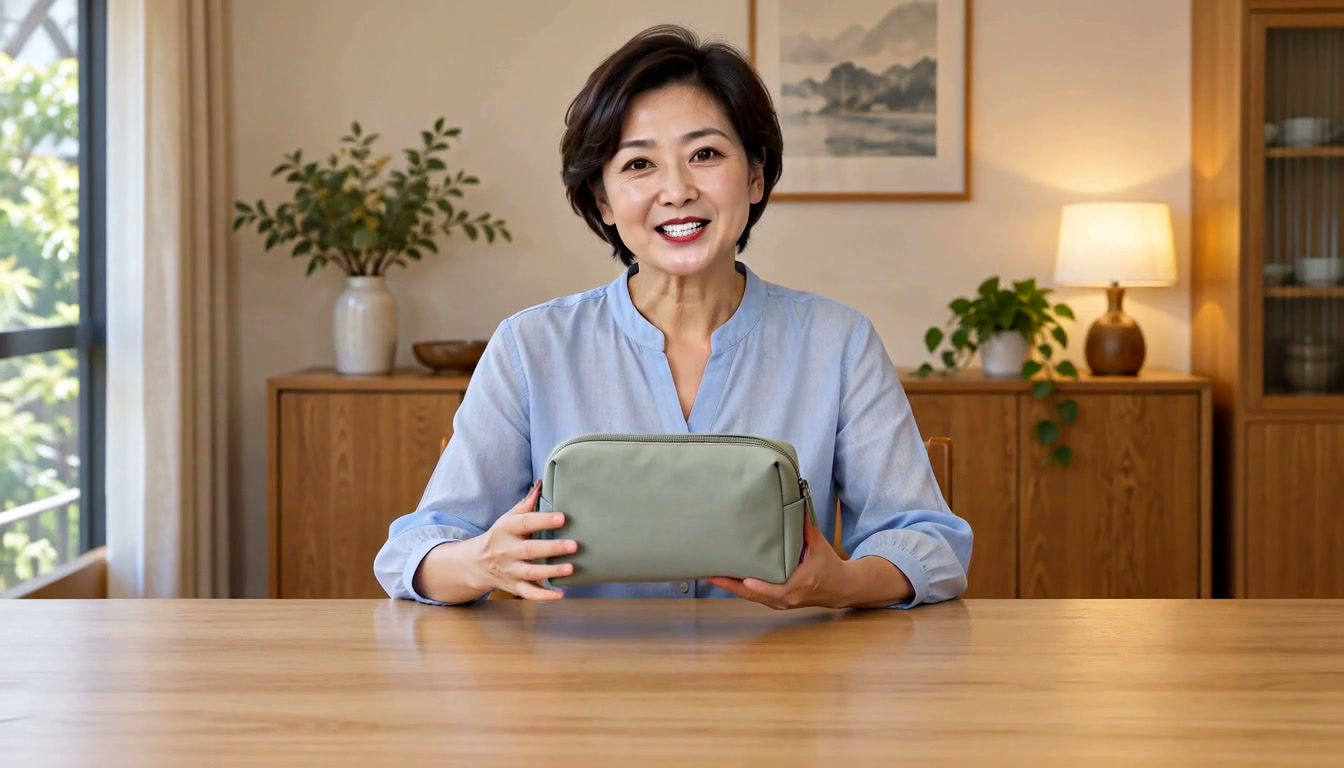}{100 126 100 0}\\[2pt]
\QualVThirteenSceneRef{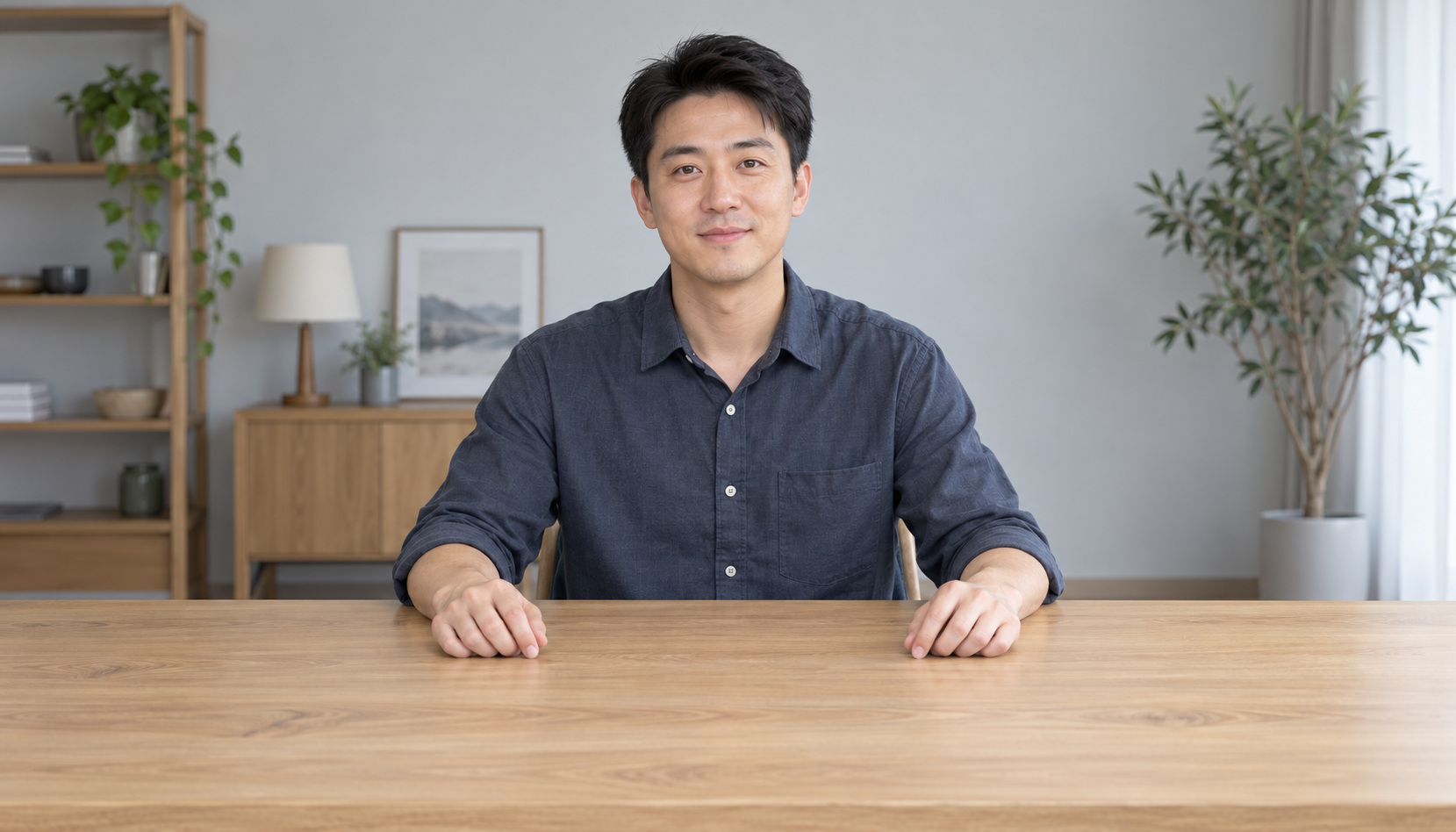}{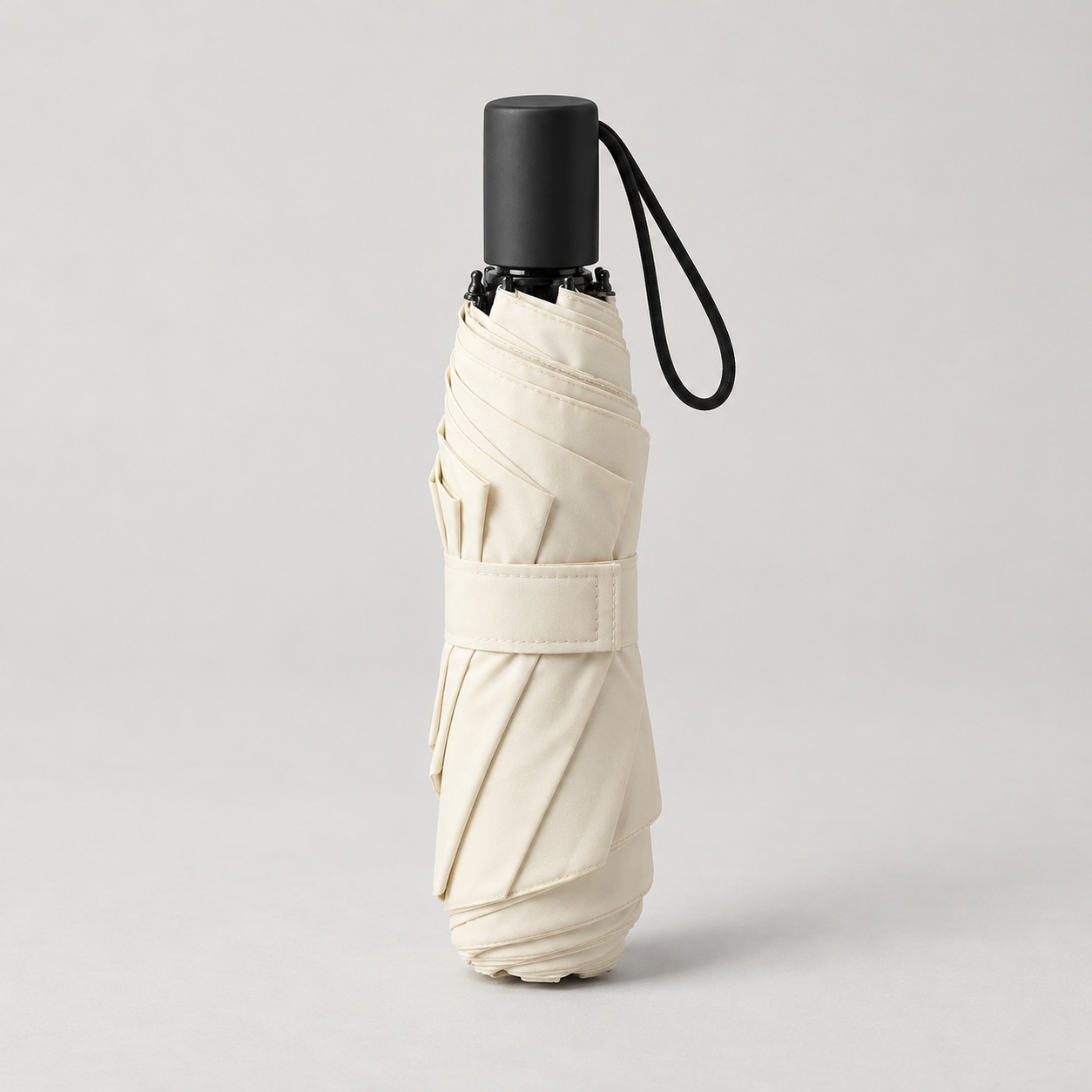}{430 100 390 80} &
\QualVThirteenFrame{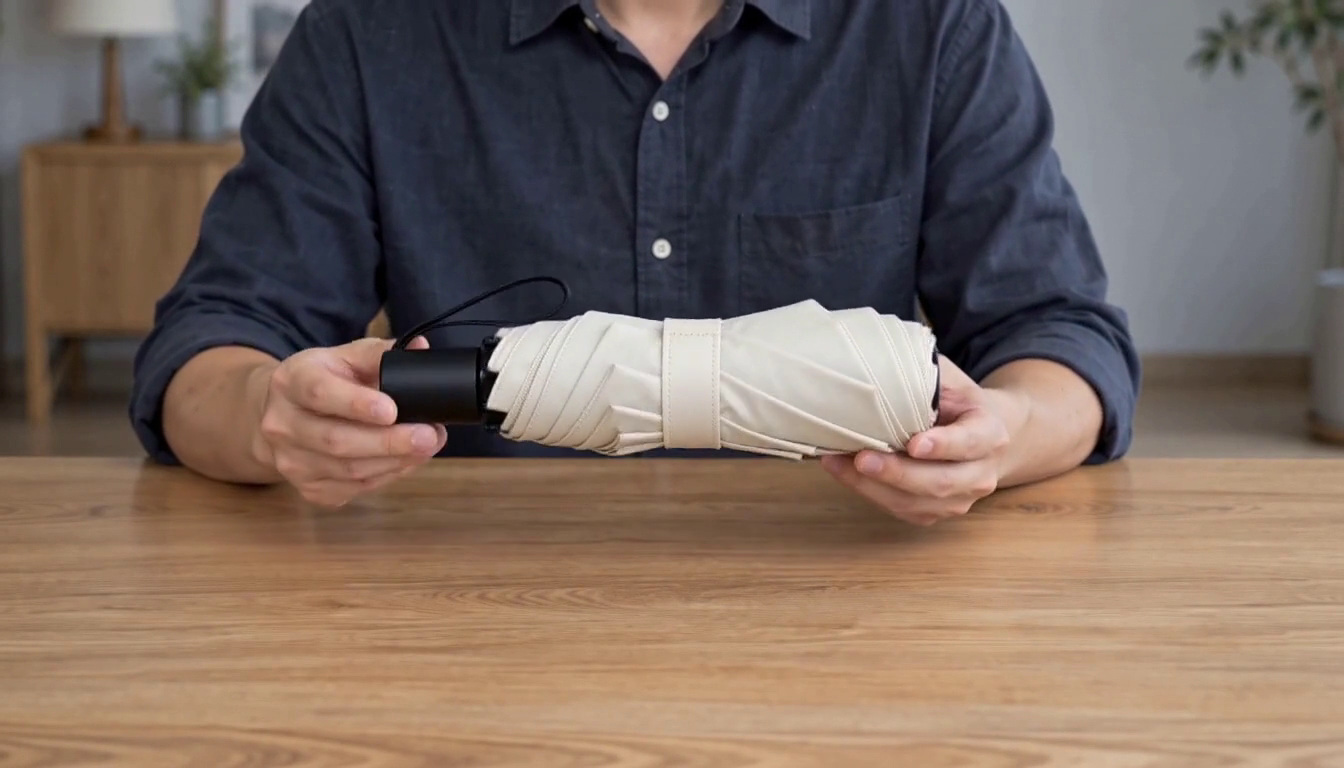}{0 6 0 6} &
\QualVThirteenFrame{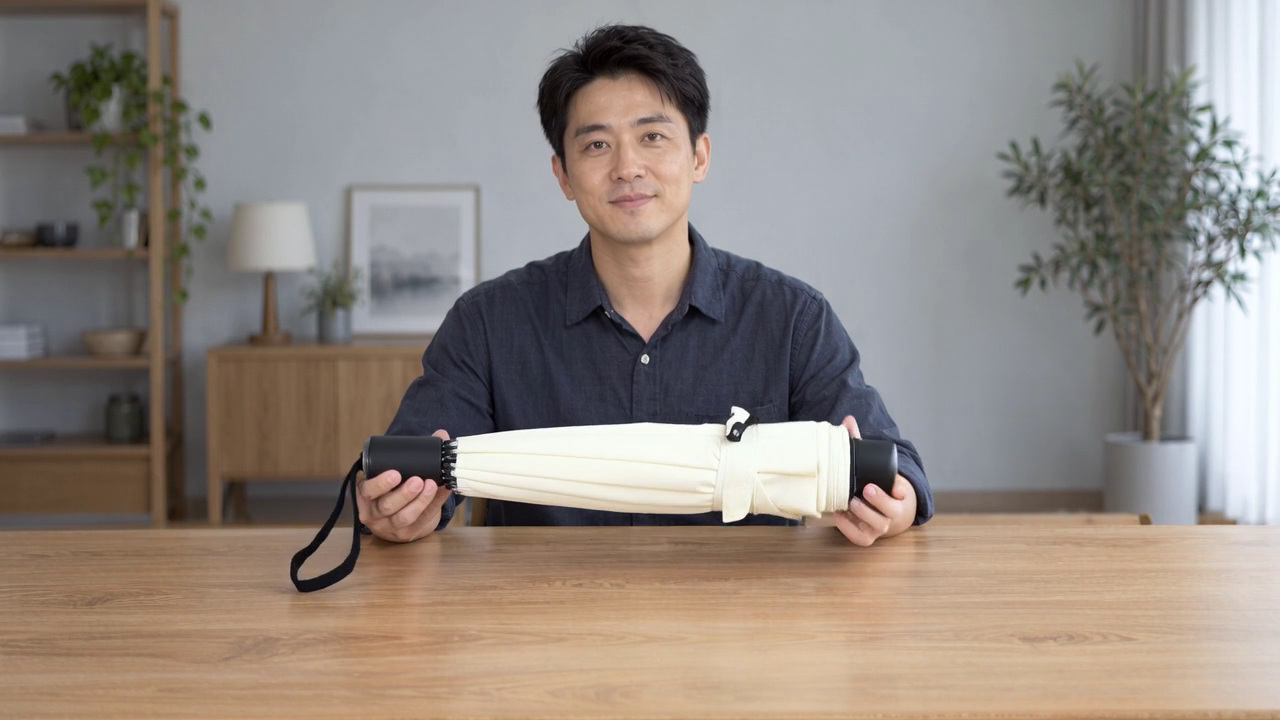}{0 0 0 0} &
\QualVThirteenFrame{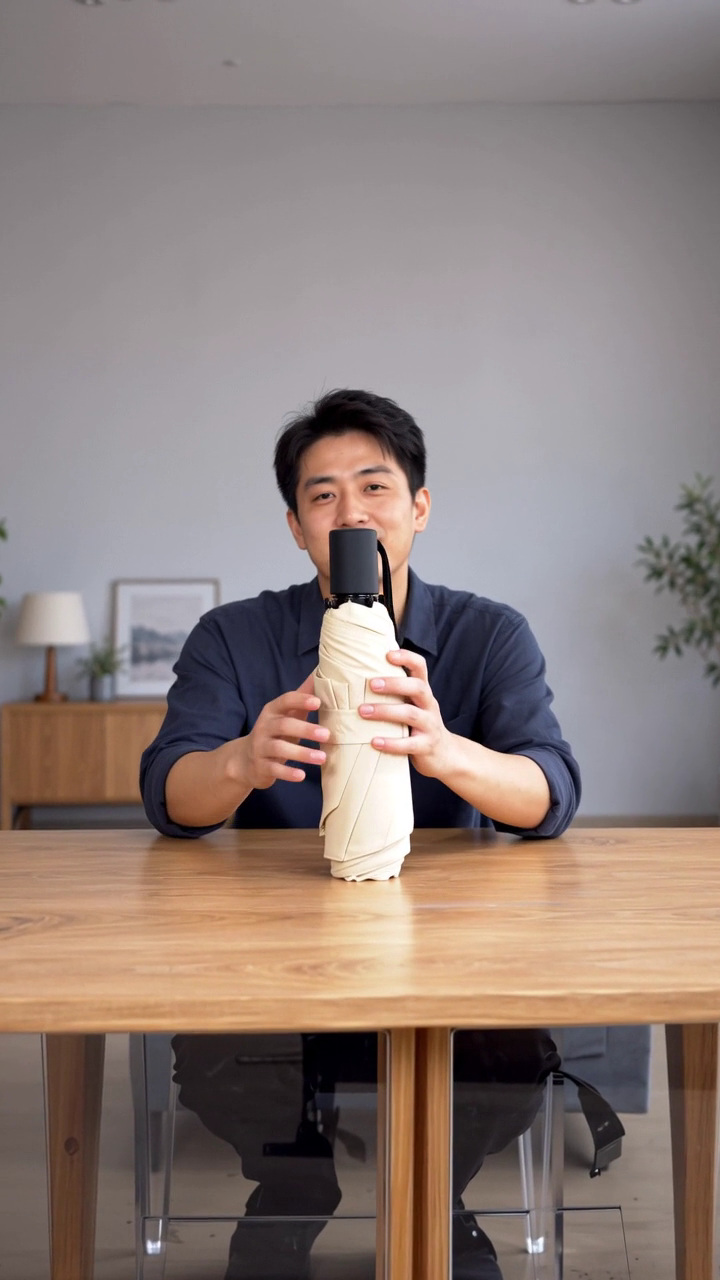}{0 365 0 510} &
\QualVThirteenFrame{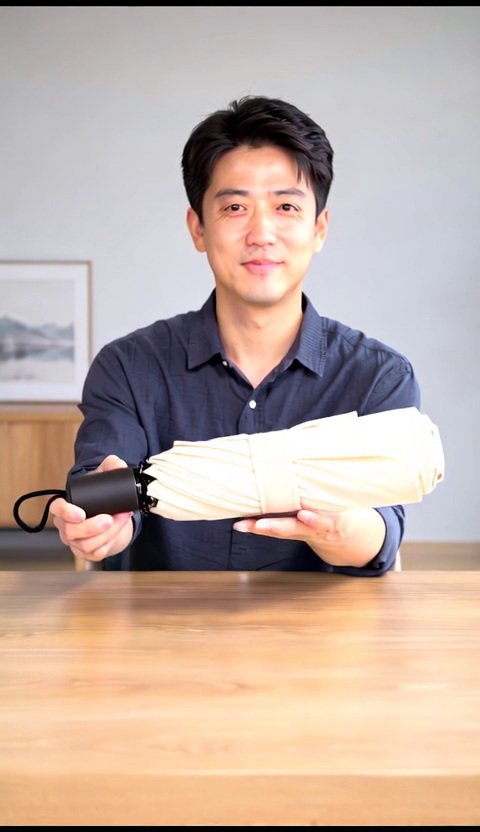}{0 242 0 320} &
\QualVThirteenFrame{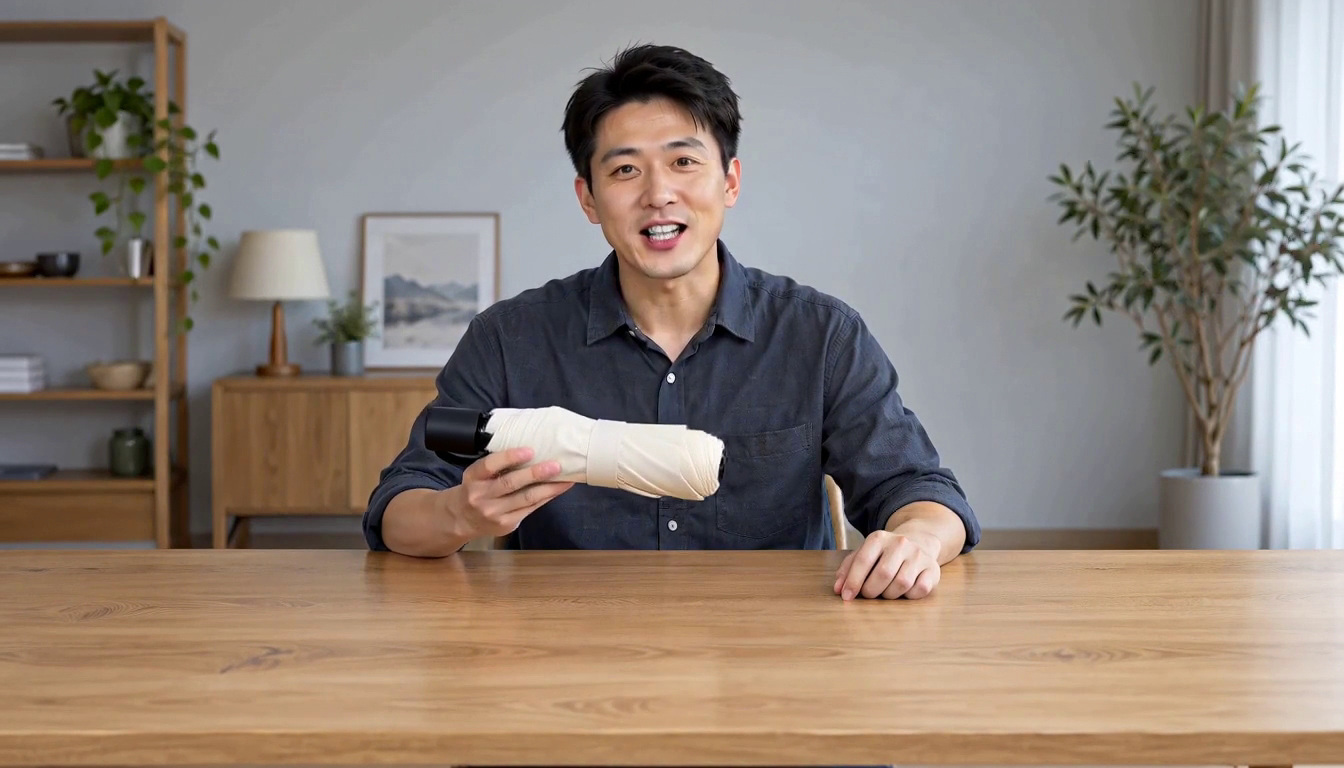}{0 6 0 6}\\
\end{tabular}
\endgroup
\caption{\textbf{Multi-reference interaction results.}
Reference inputs at left; outputs are landscape crops.
Rows: scarf adjustment (MultiOmni-board3-001, 4\,s); guitar interaction (MultiOmni-board3-003, 9\,s; SkyReels-V3/Phantom: 4.5\,s); cosmetics-bag presentation (case 008, 72\,s); and closed-umbrella presentation (case 006, 108\,s).}
\label{fig:qualitative_v10}
\end{figure*}

\begin{table}[t]
\centering
\small
\setlength{\tabcolsep}{3.6pt}
\renewcommand{\arraystretch}{1.12}
\caption{\textbf{Minute-scale continuity.} Equal-case means over ten approximately two-minute sequences. DINO$_{\rm temp}$ uses successive 1\,Hz samples; DINO$_{\rm bnd}$ and Luma compare adjacent segment-boundary frames.}
\label{tab:long_external}
\begin{tabular}{@{}lccc@{}}
\toprule
Method & DINO$_{\rm temp}\uparrow$ & DINO$_{\rm bnd}\uparrow$ & Luma $\downarrow$\\
\midrule
MiniMax-H3 & $0.832$ & $0.831$ & $10.2$\\
VACE & $0.848$ & $0.645$ & $9.12$\\
SkyReels-V3 & $0.815$ & $0.609$ & $14.3$\\
Phantom & $0.812$ & $0.589$ & $13.8$\\
\TaoMateResultLabel & $\mathbf{0.921}$ & $\mathbf{0.990}$ & $\mathbf{0.613}$\\
\bottomrule
\end{tabular}
\end{table}

\paragraph{Long-Stream System Comparison.}
\Cref{tab:long_external} compares continuous generation with independently generated instruction clips on the ten long cases.
TaoMate-R1 achieves DINO temporal similarity of 0.921, compared with 0.812--0.848 for the clip-generation systems.
At segment boundaries, DINO similarity is 0.990 versus 0.589--0.831, and the mean luminance jump is 0.613 versus 9.12--14.3.
These results compare the complete generation protocols; the same-backbone comparisons above isolate the choice of adaptation assignment.

\begin{figure}[!t]
\centering
\begin{tikzpicture}
\node[anchor=south west,inner sep=0] (timeslice) at (0,0)
  {\includegraphics[width=.76\linewidth]{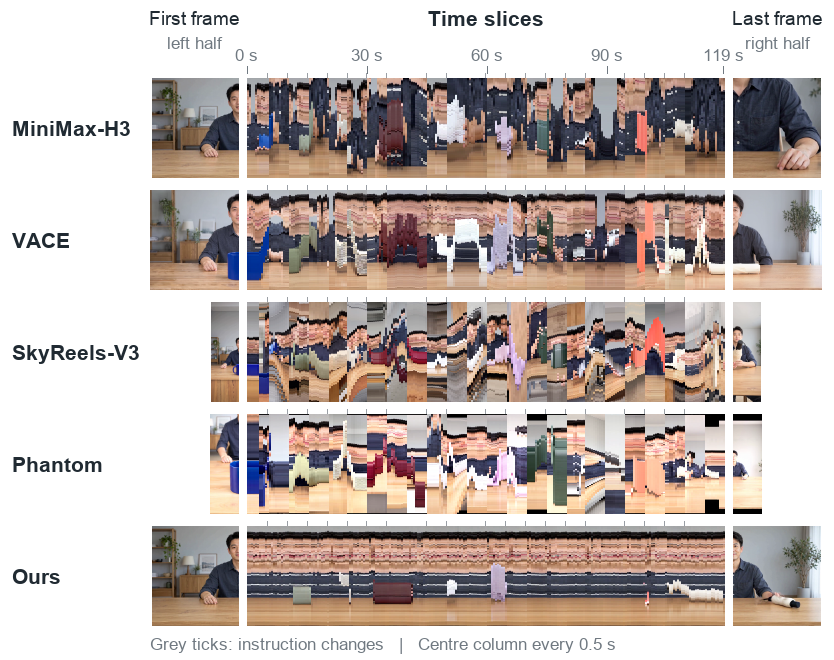}};
\begin{scope}[x={(timeslice.south east)},y={(timeslice.north west)}]
\fill[white] (.008,.095) rectangle (.174,.160);
\node[anchor=west,inner sep=0,text={rgb,255:red,32;green,42;blue,50},
  font=\sffamily\bfseries\fontsize{9.3}{11}\selectfont]
  at (.0144,.1273) {TaoMate-R1};
\end{scope}
\end{tikzpicture}
\caption{\textbf{Scene continuity over a two-minute interaction.}
Each row combines the first frame's left half, centre columns sampled every 0.5\,s, and the last frame's right half (case 006, 119\,s).
Grey ticks mark instruction changes; endpoints retain native aspect ratios.
The bands contrast composition changes in clip-wise generation with stable presenter and table geometry in the TaoMate-R1 stream.}
\label{fig:timeslice}
\end{figure}

\Cref{fig:timeslice} arranges frame columns sampled every 0.5\,s along a timeline.
Independent clip joins produce bands in head height, body scale, and background appearance, while the continuous stream maintains more stable presenter and table geometry.

The time-column representation relates these bands to changes in scene composition.
A change in camera framing or subject scale shifts several horizontal structures together, producing a discontinuity across the sampled timeline.
In the continuous stream, the table edge and background patterns trace more continuous paths through successive instructions.
This connects the boundary scores in \cref{tab:long_external} to visible scene properties such as presenter scale and background placement.
Together, the visual comparisons examine reference-conditioned interactions in \cref{fig:qualitative_v10}, localized background appearance in \cref{fig:long_range_change}, and scene-level continuity in \cref{fig:timeslice}.

\paragraph{Audiovisual Measurements.}
On the 22 native-audio cases, TaoMate-R1 has higher mean AV alignment than Operation split (4.77 versus 4.59).
Operation split has higher audio QA (0.809 versus 0.736) and lower Desync (0.282 versus 0.445), showing a different ordering for audio content and synchronization.

\FloatBarrier
\section{Conclusion}
We presented TaoMate-R1, a streaming multi-reference generator that groups denoising and context writing by scene establishment and continuation, with SGF coupling the roles through generated history.
A local gradient analysis characterizes the directions retained by this grouping and by operation-based sharing.
On the H3 backbone, TaoMate-R1 achieves higher mean CLIP and DINO temporal similarity across the three evaluated interaction settings at the same two-adapter budget.
It also reduces discontinuities at instruction boundaries in long streams.
The resulting comparison makes the specialization axis an explicit design variable in streaming multi-reference generation.

\label{paper:main_end}
\clearpage
\bibliographystyle{unsrtnat}
\bibliography{references}
\end{document}